\documentclass{article}

\usepackage[english]{babel}
\usepackage{float}
\usepackage{longtable}
\usepackage{subcaption}
\usepackage{enumitem} 
\usepackage[letterpaper,top=2cm,bottom=2cm,left=3cm,right=3cm,marginparwidth=1.75cm]{geometry}

\usepackage{amsmath}
\usepackage{graphicx}
\usepackage[colorlinks=true, allcolors=blue]{hyperref}

\title{Machine Learning-Based Classification of Vessel Types in Straits Using AIS Tracks}
\author{
  Jonatan Katz Nielsen$^{a}$ \\
  \\
  $^{a}$Copenhagen Business School, Frederiksberg, DK
}

\begin{document}
\maketitle
\begin{figure}[H]
    \centering
    \includegraphics[width=1\linewidth]{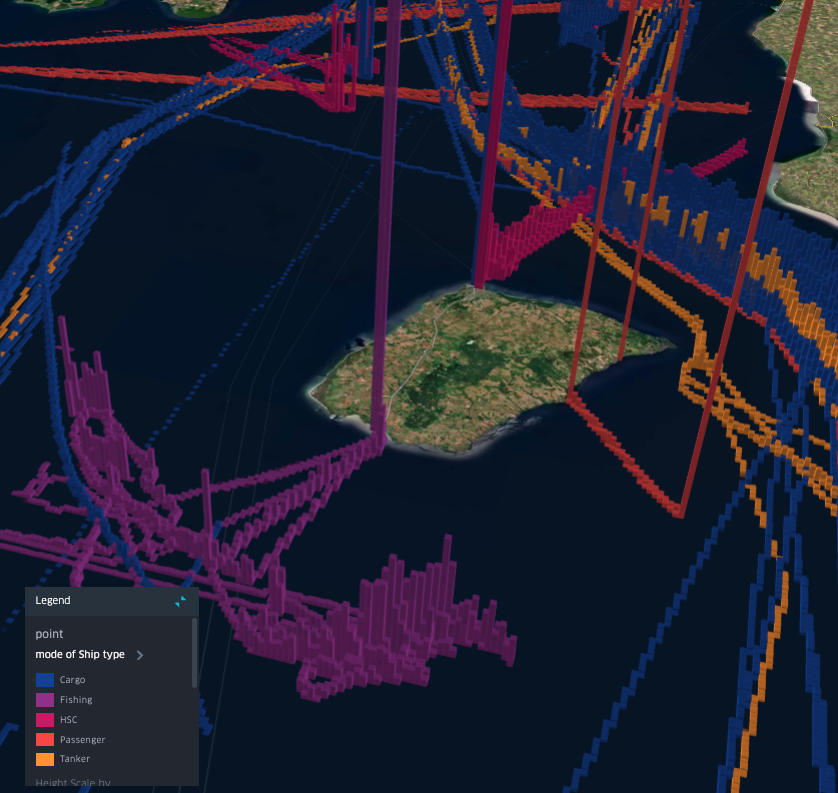}
\end{figure}

\newpage
\tableofcontents
\newpage
\section{Introduction}
\subsection{Motivation}
The world had an estimated 4.9 million fishing ships in 2022 \cite{fao2024}. This represents an overall increase compared to the early 2000s, when the global fleet was estimated at about 4 million vessels \cite{fao2002}. Along with the rapid increase in vessels, the environment surrounding maritime traffic safety has become more complex and the chance of overlooking critical events has increased. Over the past decade however, annual shipping losses have declined by 70\%, with reported losses decreasing from 89\% in 2014 to 26\% in 2023 \cite{allianz-2024}. This significant reduction, all while vessels have increased manyfold is attributed to advancements in safety measures, improved ship design, and stricter regulations \cite{allianz-2024}. Among those, systems capable of tracking and monitoring vessel activity has been one of the most crucial factors in preventing sabotage, collisions and defecting unreported ships \cite{huang-2023}. The main revolution within this area is the Automated Identification System (AIS). Traditional vessel navigation relied primarily on paper charts, radars and was largely dependent on the sailor’s sailing experience. The development of AIS was influenced by maritime incidents, notably the Exxon Valdez oil spill in 1989, which made it clear that someone had to develop better vessel tracking and communication systems \cite{wolfe-1994}. In response, the United States Congress passed the Oil Pollution Act of 1990, which included budget for the U.S. Coast Guard to develop a vessel tracking system for tankers operating in Alaskan waters \cite{kim-2002}. Early in the 2000s, it was discovered that AIS signals could be received at distances up to 400 kilometers above the Earth's surface, far exceeding the typical surface range of approximately 40 kilometers. This discovery expanded AIS from a coastal and ship-to-ship tracking application to a vessel management system with critical global coverage \cite{scully-2013}.
\newline
\newline
Illegal, unreported, and unregulated (IUU) fishing, along with other forms of maritime crime such as smuggling and unauthorized resource extraction, presents a major challenge to global ocean governance. The UN Food and Agriculture Organization (FAO) estimates that IUU fishing accounts for up to 26 million tonnes of fish annually, representing losses of over \$20 billion worldwide \cite{fao-2016}. Beyond economic impacts, such activities undermine sustainable fisheries management, damage marine ecosystems, and disadvantage coastal communities that depend on legitimate fishing \cite{sumaila-2020}. Monitoring and classifying vessel activity at scale is therefore critical. Many vessels engaged in IUU fishing or smuggling deliberately misreport their type or manipulate Automatic Identification System (AIS) data to obscure their operations \cite{taconet-2019}. Accurately distinguishing vessel types such as fishing, cargo or tankers, offers a foundation for identifying abnormal behavior patterns and prioritizing enforcement resources. This makes automated vessel classification a key enabler for addressing environmental, economic, and security concerns in maritime domains.
\newline
\newline
This project aims to develop an accurate approach for classifying vessel types to solve this problem. Success for the project is based on the classifier model's ability to deliver reliable classifications for common ship types (fishing, cargo, tankers, passenger, etc.). Other more rare types such as law enforcement, military, cruise ships would require a larger dataset specifically training for these types. It is important to note that this study only attempts to predict ship types for \textit{moving} vessels (filtering out stops longer than 1 hour). To clarify the main objective of this work, two research questions have been made: 

\subsubsection{Research questions}
\begin{itemize}
  \item Which features can be extracted from AIS data for recognizing moving vessel types?
  \item How can reasonable vessel type recognition be performed using different classification models and how can they be evaluated?
\end{itemize}

Although this research makes it possible to identify ship types based on vessel and trip features, this paper does not go into detail about how to apply these vessel types to find anomalies in this data. Had the scope of the paper allowed it, this could have potentially led to other interesting insights.

\section{Data}
The following section describes the dataset used in the study, how it is accessed, pre-processed and how features are extracted.
\subsection{Dataset and data source}

The data used for this study is historical AIS data provided by the Danish Maritime Authority \cite{dma-2024}. The time distribution is 8 days from end of 22. January 2025 to end of 30. January 2025. Specifically, the timestamps first and last timestamps are: 2025-01-22 23:59:59 to 2025-01-30 23:59:58. A dataset (.csv file \~4GB in size) is available to download every 24 hours. Each daily dataset contains approximately 20 million AIS rows, which is why the timeframe is rather short. Since ships move slow and the data need to cover longer trajectories to be able to classify movement patterns, the source data being many points is a must. All positions covering the same MMSI are converted into a single row for each trip, which means that the row on which the training is performed is much smaller. This is further covered in the preprocessing section below. 
\newline
\newline
Dataset can be accessed at: \href{https://web.ais.dk/aisdata}{https://web.ais.dk/aisdata}. All data is processed on a \textbf{Hetzner GEX130} server with 256GB ram, Intel Xeon 24 core CPU and an Nvidia RTX 6000.
\newline
\newline
Data on the web portal is delayed 48 hours from when it was received. This is more than sufficient for the purpose of training and evaluating a classification model, but when the model is deployed in practice, a real-time feed from a restricted source is used. AIS data messages are structured such that regular position updates are sent at a specific rate, often every 6 minutes. These data are also known as "dynamic" data, or message type 5 data, meaning that each position usually contains new information like speed, status, coordinates. At beginnings of journeys, sometimes at random, static vessel data is transmitted. This data contains information about the vessel size, type of cargo, call sign, ETA, position of fixed antenna on vessel and more. Before moving onto pre-processing and feature extraction, data is restricted to a specific area in the Baltic Sea. This is done with the inpoly python framework introduced by \cite{kepner-2016}. The reason for this specific area of interest, is that ships cruising for various different purposes cross this path, which include passenger trips to/from Bornholm, fishing vessels departing from the island and cargo ships as well as tankers passing through the strait from/to the Baltic Sea. The polygon bounds are visualized in figure \ref{fig:bornholm-area}, and exact coordinate bounds can be found in Appendix 1. 
\begin{figure}[H]
    \begin{subfigure}{0.5\linewidth}
        \centering
        \includegraphics[width=\linewidth]{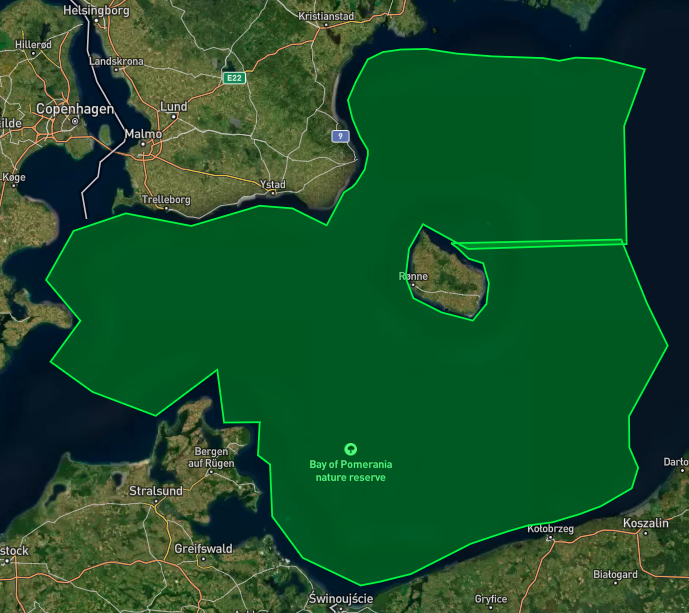}
        \caption{Baltic Sea / Bornholm Strait area of interest}
        \label{fig:bornholm-area}
    \end{subfigure}
    \begin{subfigure}{0.5\linewidth}
        \centering
        \includegraphics[width=\linewidth]{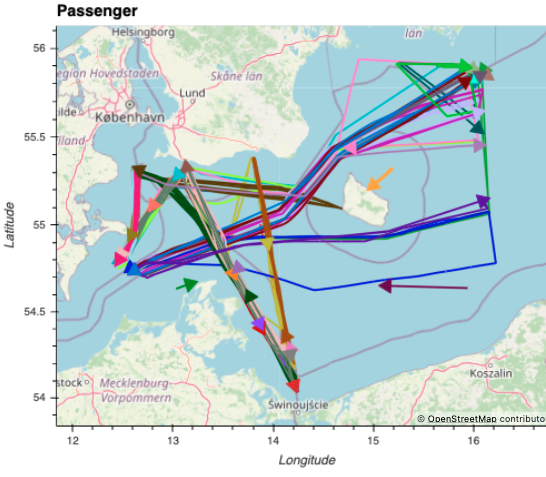}
        \caption{23. February passenger traffic in AOI}
        \label{fig:bornholmareatraffic}
    \end{subfigure}
    \caption{Combined figure showing (a) the Baltic Sea / Bornholm Strait area of interest and (b) Day's worth of traffic in AOI.} \label{fig:combined-figure}
\end{figure}
The table in Appendix 9 shows every raw property available in the dataset, and how often it is transmitted (i.e. dynamic/static).

\subsection{Pre-processing}
\begin{figure}[H]
    \centering
    \includegraphics[width=1\linewidth]{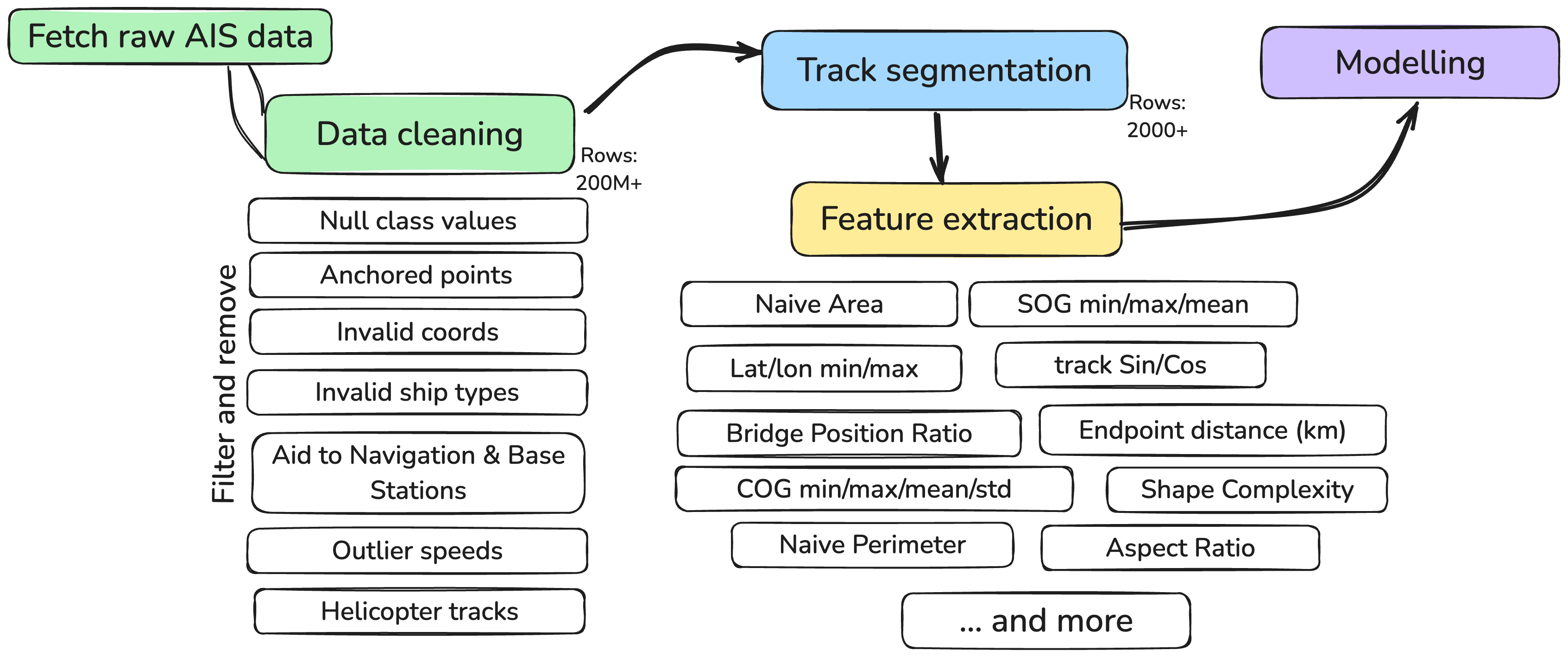}
    \caption{Overall structure of cleaning and feature extraction. Drawn with Excalidraw.com}
    \label{fig:pipeline}
\end{figure}
Which raw AIS data fields are reported is largely up to the sailor as mentioned in the introduction. Because of this, AIS data can be wrong and manipulated due to either faulty sensor readings or deliberate actions \cite{wu-2016}. Since AIS data is not suitable for performing traditional classification and machine learning techniques out of the box, it is necessary to perform data cleaning and pre-processing before moving on to creating features and building a classification model. First, duplicate rows were identified, of which there were none (since the timestamp of each AIS transmission is always unique to each MMSI, due to the way AIS transponders work \cite{marques-2019}. As stated above, the purpose of the classification task is to identify which type a specific vessel is, based on its features and trip data. Figure \ref{fig:pipeline} depicts the overall approach to how the data cleaning and feature extraction is done in this study.  Ships continue transmitting AIS data even when they are anchored, drifting, moored or in a holding pattern, which is why data cleaning must be performed on the dataset before continuing to feature extraction and creating a model. 
\begin{figure}[H]
    \centering
    \begin{subfigure}{0.42\linewidth}
        \centering
        \includegraphics[width=\linewidth]{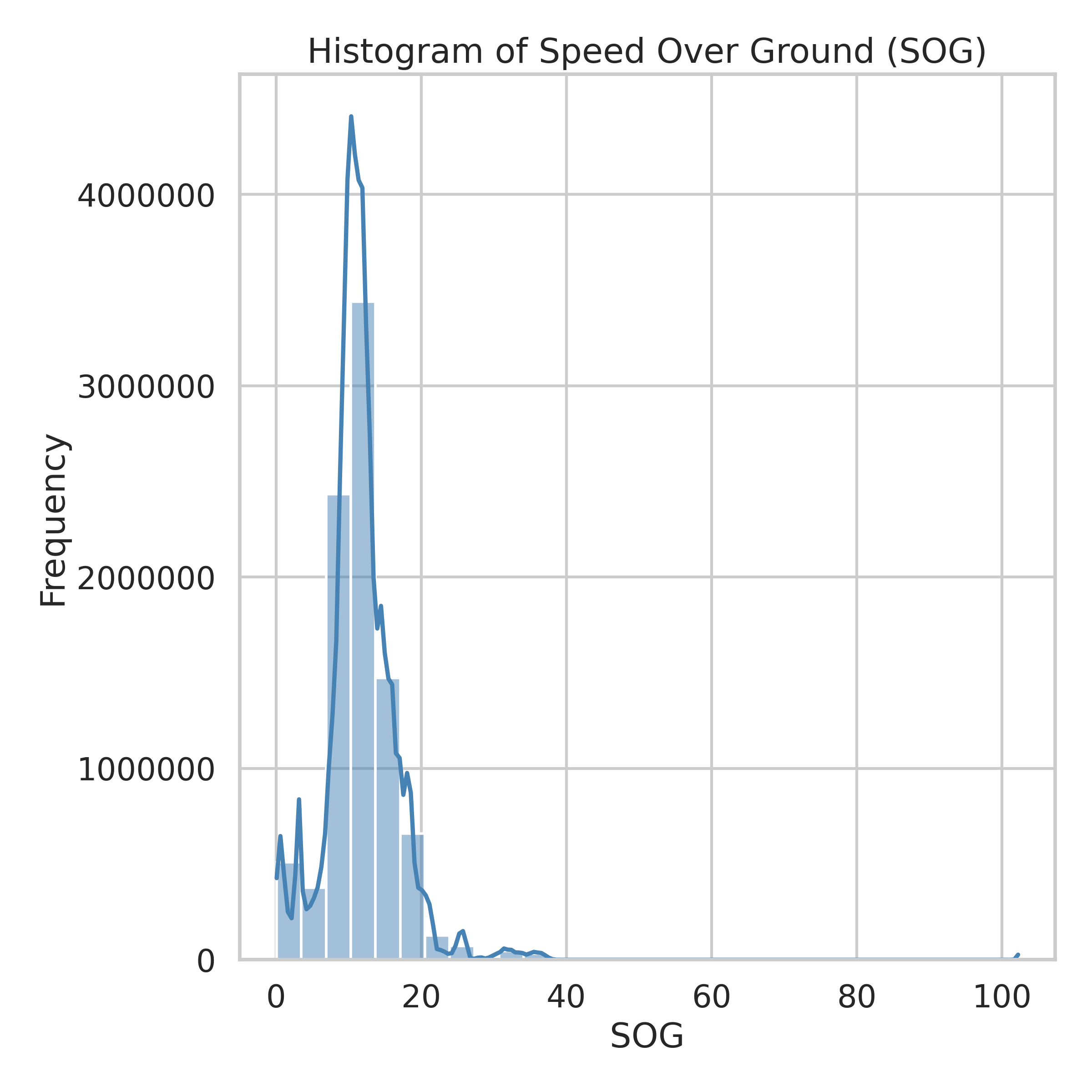}
        \caption{Histogram of Speed over Ground in dataset}
        \label{fig:speed-histogram}
    \end{subfigure}
    \hfill
    \begin{subfigure}{0.42\linewidth}
        \centering
        \includegraphics[width=\linewidth]{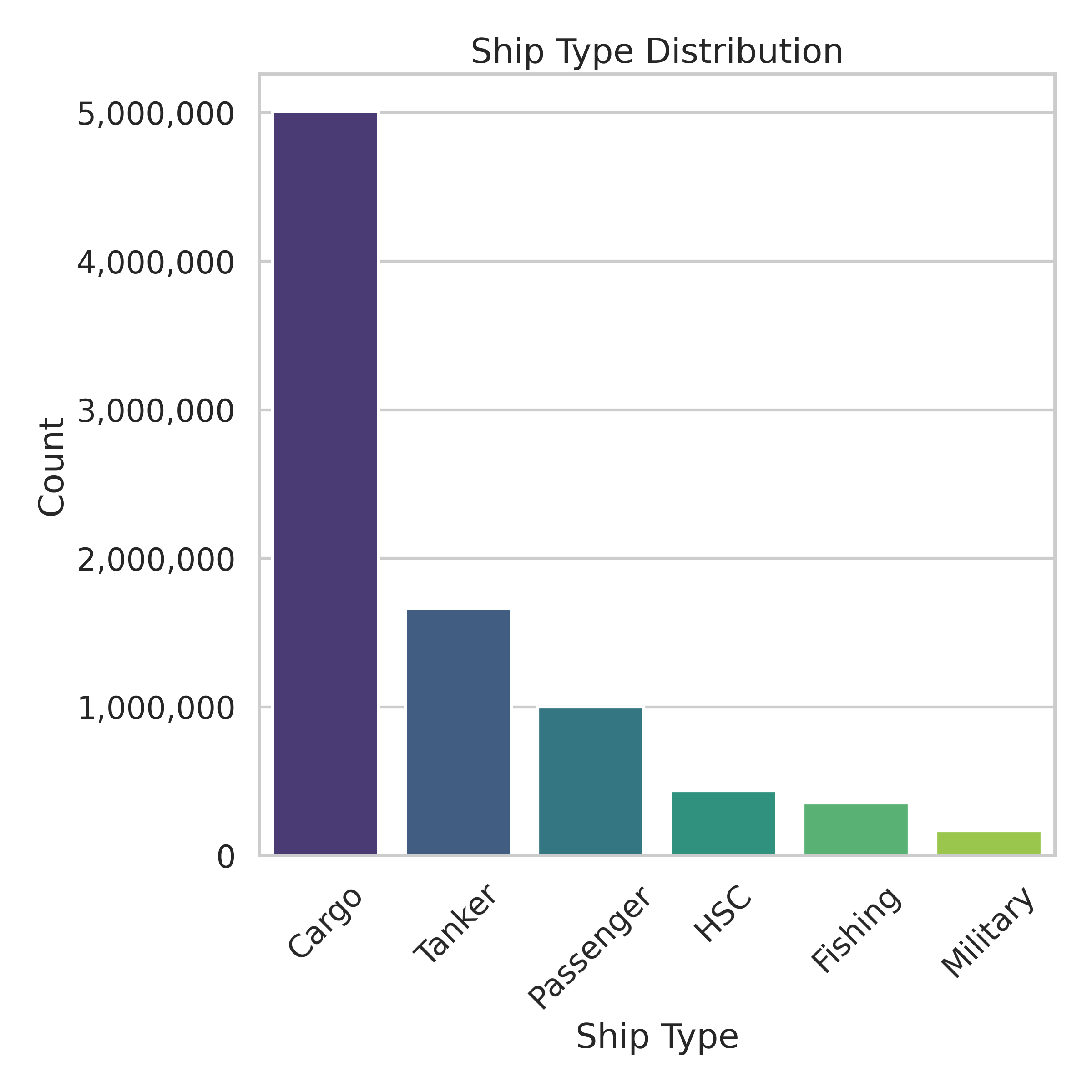}
        \caption{Most common ship types in dataset}
        \label{fig:ship-distribution}
    \end{subfigure}
    \caption{(a) Histogram of Speed over Ground and (b) Distribution of most common ship types in the raw dataset.}
    \label{fig:dataset-analysis}
\end{figure}
Plotting the SOG on a histogram shows speed outliers (figure \ref{fig:speed-histogram}) (note that this is after cleaning 0 speed transmissions, which is often wind-farms and anchored ships). By manually inspecting a few of the SoG values over 80, it becomes clear that it is wrongful transmissions and not really fast vessels. These outliers are removed, along with helicopter tracks, and other noise not covering actual vessels (\textit{2\_cleaning.ipynb}). In figure \ref{fig:dataset-analysis}, the ship type distribution of a single day's AIS traffic is plotted. By plotting weekdays vs. weekends it becomes clear that overall the distribution of ship types looks similar day by day, slightly varying with more fishing vessels seen during weekdays and more pleasure vessels seen during Weeekends. This is relevant to consider, when picking the time frame. For this study, the time frame is 8 days starting on a Wednesday, which means that only one weekend is present in the dataset. By visualizing the classes, a clear imbalance is visible in the dataset, which will need to be handled in the modeling section when fitting models on the dataset. Additionally, AIS messages with Navigational Status set to "\textit{Moored}", "\textit{Anchored}", "\textit{Vessel constrained by her draught}". This cleaning of static ships in ports are on purpose, because the purpose of the paper as described in the research question is to predict ship types for \textit{moving} vessels. 

\subsubsection{Forward/backward filling static voyage data}
Since dynamic AIS messages are transmitted much more often than static voyage data, it is necessary to do some forward and backwards filling of missing values \cite{ferreira-2023}. By applying merge\_asof in pandas, every dynamic AIS message that is sent after a static voyage message has its missing values filled with the data from that previous row. For the ones that's missing, often because the voyage message came just before the start of the study's time range, it is filled backwards to find the closest voyage information to the dynamic AIS message. This ensures that each AIS record has the correct ETA, Destination, Cargo type, properties etc.

\subsubsection{Location outliers}
By visualizing the Latitude and Longitude columns in a scatter plot, clear deviations are visible. Although the point in polygon algorithm from \cite{kepner-2016} cleaned most values as mentioned above, it resulted errors on invalid latitude/longitude pairs. These are further cleaned by drawing a bounding box around Denmark and filtering all outlier values. By manually inspecting the values, it is clear that these values are faulty readings, possibly caused by GPS jamming in the baltic sea with HMI \cite{siegert-2017}. The bounding box used is: [4.25, 53.61, 19.54, 61.89] (EPSG:4326)
\begin{figure}[H]
    \centering
    \begin{subfigure}{0.45\linewidth}
        \centering
        \includegraphics[width=\linewidth]{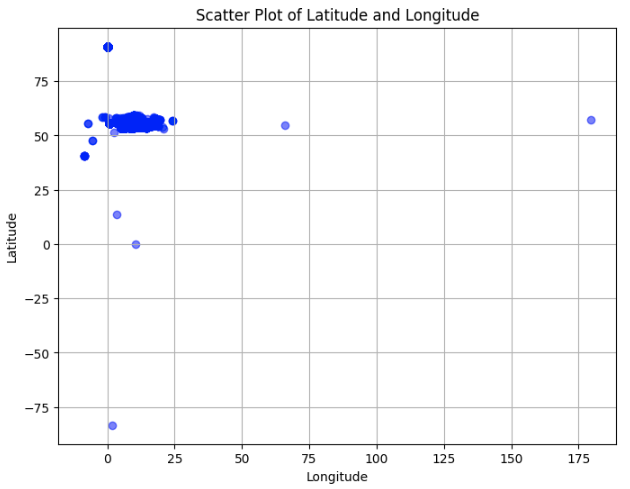}
        \caption{Location outliers}
        \label{fig:locoutliers}
    \end{subfigure}
    \hfill
    \begin{subfigure}{0.45\linewidth}
        \centering
        \includegraphics[width=\linewidth]{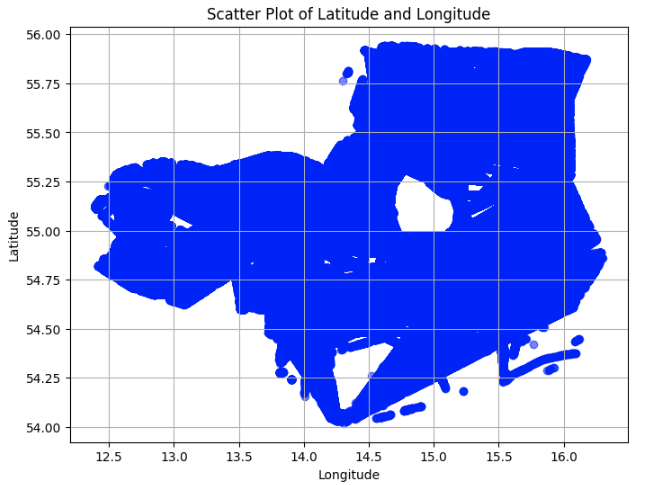}
        \caption{Scatterplot after cleaning}
        \label{fig:scatteraftercleaning}
    \end{subfigure}
    \caption{Comparison of location outliers before and after cleaning}
    \label{fig:comparison2}
\end{figure}

\subsection{Vessel trajectory segmentation}
\begin{figure}[H]
    \begin{subfigure}[b]{0.5\linewidth}
        \centering
        \includegraphics[width=\linewidth]{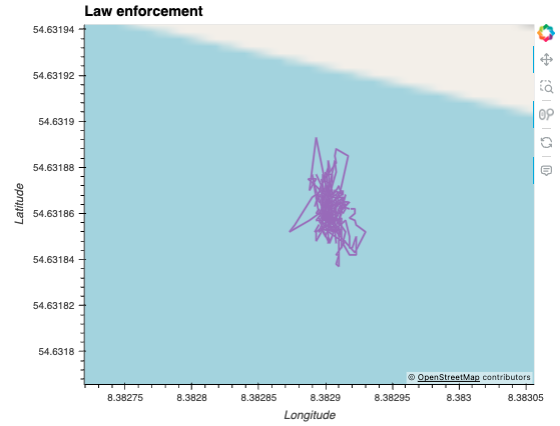}
        \caption{Anchored ship (MMSI: 219024604) with navigation status "\textit{Under way using Engine}"}
        \label{fig:anchored-moored-loitering}
    \end{subfigure}
    \begin{subfigure}[b]{0.5\linewidth}
        \centering
        \includegraphics[width=\linewidth]{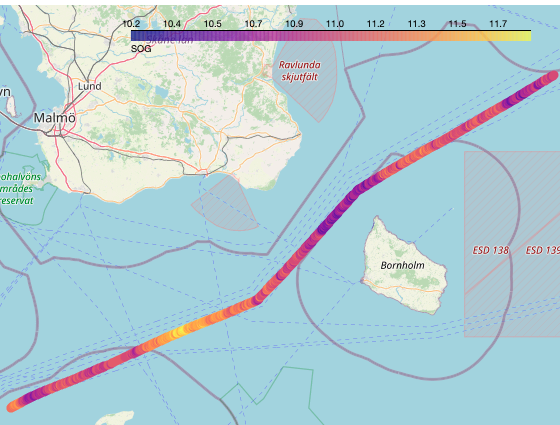}
        \caption{Moving trajectory of a ship (MMSI: 209662000) with nav status "\textit{Under way using Engine}" - legend indicates SoG (speed)}
        \label{fig:trajcorrect}
    \end{subfigure}
    \caption{Comparison of ship trajectories. (a) Raw anchored ship positions. (b) Moving trajectory of ship.}
    \label{fig:combined-trajectory}
\end{figure}

After obtaining and cleaning the individual AIS messages in the dataset, the next step is to generate trajectories based on the raw AIS messages for each ship. A trajectory in this sense can be defined as consecutive AIS positions (coordinate pairs) for the same ship (i.e. with the same MMSI) with no longer stops. Figure \ref{fig:trajcorrect} shows how a track looks after being correctly segmented into a trajectory. This correctly segments it into one trajectory from when it goes into the area of interest until it goes out of the area of interest, the same trajectory, although reversed, can be seen 6 days later when the same MMSI comes back - had the trajectory segmentation not been performed, these points would be segmented into one, which is not useful for the purpose of the classification task. While calculating trajectories, many features are extracted. This is expanded upon in the section below. Although AIS positions ideally report when ships are moored or anchored, it is clear by visualizing positions for a number of ships that this is not always the case. The navigational status of this law enforcement ship named \textbf{LUNDEN} (\textit{MMSI: 219024604}) is set to "\textit{Under way using Engine}", when in fact it has been moored at the same port for over 48 hours. It is necessary to filter out this noise in the dataset, so that only ships moving are used for the classification model. This is done by segmenting a track if a ship is the same place without moving for 1 hr+. Afterwards, feature extractions are used to further clean still ships, this is done by cleaning trajectories where trajectory\_length\_km (described below) are less than 200 meters.

The total amount of trajectories is shown in table \ref{tab:trajs} below. An alternative approach could be to use an unsupervised spatial clustering algorithm like DBSCAN to find stationary clusters. These clusters would indicate vessel stops in ports and at sea. This point is further expanded on in the discussion chapter.
\begin{table}[H]
\centering
\begin{tabular}{c|ccllccllc}
\textbf{Ship type}& Cargo & Fishing &HSC &Dreging& Passenger& Tanker &Military &Police &Total\\\hline
\textbf{Quantity}& 999& 60 &100 &27& 417& 336 &28 &26&1910\\\end{tabular}
\caption{\label{tab:trajs}Number of created trajectories by the different ship types}
\end{table}
\subsection{Feature engineering}
Selecting the right features are critical to archiving accurate performance in a classification model \cite{provost-2013}.  Modern transponders and digital sensors enable the collection of a large amount of data from objects in motion. AIS is no exception to this. Based on the data we have acquired, many different features can be created and extracted to uncover various hidden patterns and use them to solve the problem of ship classification. Other papers have already introduced various features for identifying vessel types from static vessel reference data features \cite{huang-2023}. These include both geometric and motion behavior features.  The overall categories of features can be described as: \textit{vessel shape features}, \textit{temporal features}, \textit{geospatial features} and \textit{kinematic features}.

\subsubsection{Kinematic features}
For the kinematic movement based features, the Speed over Ground property (SOG) is first split into min, max, mean, median and standard deviation. Afterwards, the same is done for calculated values for Corse over Ground (COG) (e.g. the actual direction the ship is moving). The initial course angle cosine and sine values are calculated as well. 

\subsubsection{Temporal features}
Trip start time and trip end time are properties created when building the trajectory, this makes it possible to simply calculate the duration of the trip in seconds "\textit{Trip duration in seconds}". Number of points is also listed as a property, since larger ships with more GPS types can transmit more often \cite{marques-2019}. To further utilize this fact, a property, number of positions per minute could potentially be extracted, this is not done for this specific dataset, since the trajectories are deemed too short to be valuable. 

\subsubsection{Geospatial features}
First, every AIS message is considered in the trajectory in order to build max and min latitude, longitude pairs. This makes it possible to calculate the total km2 area of the bounding box. The Haversine formula is used to calculate the great-circle distance between two points on a sphere given their latitude and longitude \cite{wikipedia-contributors-2024}. Since we have a long sequence of GPS coordinates \( (\text{lat}_i, \text{lon}_i) \), the total travel distance for the ship be calculated by summing up the Haversine distances between consecutive points: \((lat_1, lon_1), (lat_2, lon_2), \dots, (lat_n, lon_n) \)
The total travel distance is computed by summing the Haversine distances between consecutive points:
\[
D = \sum_{i=1}^{n-1} d_i
\]
where each segment distance \( d_i \) is given by:
\[
d_i = R \cdot 2 \cdot \text{atan2} \left( \sqrt{a_i}, \sqrt{1-a_i} \right)
\]
where:
\[a_i = \sin^2\left(\frac{\Delta \phi_i}{2}\right) + \cos(\phi_i) \cdot \cos(\phi_{i+1}) \cdot \sin^2\left(\frac{\Delta \lambda_i}{2}\right)\]
\[\Delta \phi_i = \phi_{i+1} - \phi_i, \quad \Delta \lambda_i = \lambda_{i+1} - \lambda_i\]
with:
\begin{itemize}
    \item \( R \) as the Earth's radius (\( R \approx 6371 \) km),
    \item \( \phi_i, \phi_{i+1} \) as latitudes in radians,
    \item \( \lambda_i, \lambda_{i+1} \) as longitudes in radians.
\end{itemize}
This feature is calculated and named "\textit{trajectory\_distance\_km}". The same approach is used for "\textit{endpoint\_distance\_km} except it only calculates start and end position, and the Haversine distance between those two points. Furthermore, this feature is used to clean stationary vessels from the trajectory list, by filtering low values of these two properties.
\subsubsection{Vessel shape features}
As shown in table \ref{tab:maritime_data_fields}, four properties are present called "\textit{A, B, C and D}". These shows the reference point of the AIS transponder on the ship. The data is only missing in 3\% of rows, and is therefore used extensively in the model.
\begin{figure}[H]
    \centering
    \includegraphics[width=0.5\linewidth]{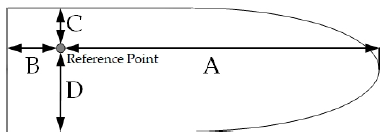}
    \caption{Reference point ABCD}  \cite{huang-2023}
    \label{fig:refpoint}
\end{figure}
By using clever equations on reference points available in the AIS data, features can be built that uncover hidden details for vessels. Since \( A \) covers from bow to the reference point of the AIS transponder, \( B \) covers stern to the reference point, \( C \) covers port to the reference point, and \( D \) covers starboard to the reference point, the vessel’s length and width are given by \( L = A + B \) and \( W = C + D \). Width and length are calculated and used as features by itself. Additionally, the aspect ratio of the ship is defined as \( \text{Aspect Ratio} = \frac{W}{L} \) and calculated as features for the vessel. The naive perimeter is given by \( \text{Naive Perimeter} = 2 \times (L + W) \), while the naive area is calculated as \( \text{Naive Area} = L \times W \). Shape complexity, which provides an indication of structural proportions, is defined as: \[ \text{Shape Complexity} = \frac{(L+W)^2}{L \times W} \]
The Bridge Position Ratio was originally introduced in \cite{huang-2023}. Since most antenna positions are located on the bridge, this equation helps identify larger container ships which adopts a double bridge, and therefore the reference point is set on the front bridge, which is different from traditional container ships, helping to identify large cargo ships vs tankers. The equation for this feature is given by:
\begin{equation}
Brige\;Position \;Ratio = \frac{A}{L}
\end{equation}
and aids in distinguishing specific vessels from others, where the length and width along with other properties are otherwise similar \cite{gcaptain-2020}.
\subsubsection{List of features}
A table with every feature along with their datatype and description can be found in Appendix 11. In total, 31 features are extracted for each trajectory.

\subsection{Preparing for applying classification models}
Finally, the AIS has been segmented into trajectories, each showing a vessel "\textit{trip}", along with features that explain geospatial, temporal, kinetic and vessel specific properties of the trip of the vessel. Before classifying the model, a couple of steps are taken:

\begin{itemize}
    \item Ship types are filtered to relevant categories (Passenger, Cargo, Tanker, Fishing, HSC).
    \item Unnecessary columns such as \textit{Name}, \textit{Callsign}, \textit{Destination}, \textit{ETA}, and \textit{MMSI} are dropped. These properties are removed since the purpose of the model is to predict ship types based on trajectory and vessel data.
    \item Categorical columns such as cargo type are encoded using LabelEncoder.
    \item Rows with missing data (10 rows) are dropped.
    \item Feature scaling is performed by standardizing numerical features using StandardScaler to ensure all features contribute equally to the model.
\end{itemize}

\subsection{Train/test split}
A usual approach to creating the train/test data split is simply splitting the dataset 80\%/20\% randomly or when working with a time series set it can be split by a specific set time \cite{provost-2013}. However, in the context of maritime trajectory prediction, where a single Maritime Mobile Service Identity (MMSI) can be associated with multiple trips or trajectories over different time periods, a naive random split can result in data leakage. Specifically, if trajectories from the same vessel appear in both the training and test sets, the model may exploit target vessel specific patterns rather than generalizable motion behaviors, leading to overly optimistic performance estimates. To mitigate this issue, a grouped splitting strategy has been applied, taking the usual approximately 80/20 split, but doing it in such a way that the same MMSI does not appear in both the training and test set, ensuring that trajectories from the same vessel are contained entirely within either the training or test set. After splitting, the dataset can be found in table \ref{tab:ship_type_distribution_train} and \ref{tab:ship_type_distribution_test}:
\begin{table}[H]
    \begin{minipage}{0.5\textwidth}
        \centering
        \begin{tabular}{c|c}
            \textbf{Ship Type} & \textbf{Count} \\
            \hline
            Cargo& 799\\
            Passenger& 334\\
            Tanker& 267\\
            HSC& 80\\
            Fishing& 48\\
            \hline
            Total&1528\\
        \end{tabular}
        \caption{Train set class distibrution of vessel types}
        \label{tab:ship_type_distribution_train}
    \end{minipage}
    \begin{minipage}{0.5\textwidth}
        \centering
        \begin{tabular}{c|c}
            \textbf{Ship Type} & \textbf{Count} \\
            \hline
            Cargo& 200 \\
            Passenger& 83 \\
            Tanker& 67 \\
            HSC& 20 \\
            Fishing& 12 \\
            \hline
            Total&382\\
        \end{tabular}
        \caption{Test set class dist of vessel types}
        \label{tab:ship_type_distribution_test}
    \end{minipage}
\end{table}
\section{Model}
Considering ship types were available in the dataset, therefore becoming labeled training data, supervised learning was solely applied in this case. Supervised learning is when a model learns from input-output pairs to make predictions on new, unseen data \cite{muller-2016}. The task specifically is classification, and the target variable is the ship type. Within machine learning, especially in supervised learning, no single algorithm can consistently excel in every problem, as its performance is dependent on data structure and features \cite{provost-2013}. Therefore, it becomes imperative to consider various models and algorithms for a given problem and evaluate their performance using train/test splits to select the best performing one \cite{suthaharan-2015}. In this paper, to find the best fitting classification algorithm on the features, five different classifiers have been experimented with, being Gaussian Naive Bayes (\textit{GaussianNB}), Support Vector Machine (\textit{SVM}) Random Forest (\textit{RF}), Decision Trees (\textit{DT}). Standard classifier evaluation metrics such as Accuracy, Precision, Recall and F-Measures has been used to measure the model’s effectiveness in classifying the vessel types. Accuracy, average precision, average recall and average F1 scores are found for each classifier, and results are shown in table \ref{tab:classifier_performance}. As can be seen in table \ref{tab:ship_type_distribution_train}  and \ref{tab:ship_type_distribution_test}, both the train and test set has clear visible class imbalance. For example, the Cargo ships (label 0) has many more samples than the Fishing class (label 1). This imbalance can affect the classifiers' ability to learn minority classes, which is why a SMOTE approach \cite{chawla-2002} is tested for every model and compared to non-SMOTE accuracy.

\subsection{Classification evaluation}
When evaluating classifiers, accuracy, precision, recall, and F1 scores are often used \cite{provost-2013}. TP, TN, FP, and FN refer to the number of true-positive, true-negative, false-positive, and false-negative samples for each class. Accuracy refers to the number of correct judgments, TP represents a judgment that positive samples are indeed positive, and TN represents the judgment that negative samples are indeed negative. FP denotes instances labeled as positive by mistake, and FN indicates positive cases that were not found. Precision \( \frac{TP}{TP + FP} \) measures the correctness of positive predictions, while recall \( \frac{TP}{TP + FN} \) reflects the ability to capture all actual positives (\textit{Appendix 7}). ROC/AUC is also used here. The ROC curve (Receiver Operating Characteristic) plots the true positive rate against the false positive rate over a range of thresholds. The AUC (Area Under the Curve) provides a single-number summary of this curve. For multi‐class problems (like ship type prediction here), one common approach is to compute ROC AUC in a one‐vs‐rest manner. This is done for the models. 
\newline
\newline
Cross validation is a method used to evaluate a models overall performance \cite{muller-2016}. A common method used is k-fold cross validation, in which the train sample is split into k subsamples, a single subsample is then retrained as data for the validation model, and the remaining k-1 samples are used for training data. The cross val is repeated k times, once for each subsample, and the results are averaged k times. A 5-fold cross validation is often used. This means that we split the data once (described above) in train/test, and then strictly use the train set with cross validation to tune and pick models. The cross validation is a stratified k fold, which ensures that each fold maintains a similar class distribution, which is important for imbalanced data, further described in \textit{Appendix 10}. Afterwards, all selected models are evaluated on the test set, to see how it would perform in a real world scenario.

\subsection{Gaussian Naive Bayes}
Gaussian Naive Bayes (GaussianNB) is a probabilistic classifier based on Bayes' Theorem assuming normally distributed features \cite{muller-2016}. It is "naive" in the sense that it assumes all features are independent given the class. It further assumes that the continuous features follow a normal (Gaussian) distribution. This will be used as a baseline model since it is one of the simplest classification algorithms. This makes it a good starting point. The target variable is "Ship type". 
\newline
GaussianNB achieves an average accuracy of \textbf{75.13\%} across the cross-validation folds, with a standard deviation of \textbf{3.34\%}. On average, when the GaussianNB model is trained on a portion of the training data and then tested on unseen data via. cross-validation, it correctly classifies about 75 out of 100 ship instances. The standard deviation of \textbf{± 3.34\%} shows that performance is fairly consistent across the different splits of the dataset. In other words, the performance of the model doesn't fluctuate wildly depending on which subset of the data it was trained or tested on. Since GaussianNB is designed to be a simple classifier \cite{muller-2016}, no hyperparameter tuning is done for the GaussianNB model, the only hyperparameter might be var smoothing, but is not regarded as being important enough to tune. The Gaussian Naive Bayes model serves as somewhat of a baseline model that other more advanced models can be compared against. The model is tested and compared with the rest of the models on the test set in the Results chapter and shown in table \ref{tab:classifier_performance}.

\subsection{Support Vector Machine}
Here a Support Vector Machine is fitted with default parameters. The default parameters is the regularization parameter C set to 1.0 and gamma set to "scale" which automatically scales the kernel coefficient based on the number of features \cite{suthaharan-2015}. Afterwards it is evaluated on the accuracy on cross validation with 5 folds. Compared to the GNB classifier, it is a more powerful model, that works by finding a hyperplane that best separates the classes in a high-dimensional space \cite{suthaharan-2015}. With a non-linear kernel, SVM can capture non-linear relationships between various features. SVMs are more computationally intensive than GNB, and require some tuning of hyperparams (like C and gamma) to archive optimal performance.
\newline
\newline
For the default SVM classifier with default parameters, an average accuracy of \textbf{82.46\%} was found with a standard deviation of \textbf{± 1.63\%}, the accuracy is  \textbf{9.76\%} better than GNB, which is a substantial improvement on the training data over the previous model when predicting vessel types.

\subsubsection{Support Vector Machine with hyperparameter tuning}
To further improve the performance of the SVM,  hyperparameter tuning is performed using GridSearchCV. For each systematic combination of parameters, it uses the cross-validation with stratified k fold (5) to evaluate the model based on accuracy. The parameter grid for the SVM includes:
\newline
\newline
\textbf{C: }The regularization parameter, which controls the trade-off between achieving a low training error and a low testing error. Smaller values of C create a smoother decision boundary, while larger values allow the model to fit the training data more closely. Default parameter is 1.0
\newline
\newline
\textbf{Kernel: }The type of kernel used. Different kernels can capture different types of relationships in the data. In the gridsearch we search for linear/RBF. The default value is RBF.
\newline
\newline
\textbf{Gamma}: Defines how far the influence of a single training example reaches. A lower value means “far” and a higher value means “close.” In our grid, we consider both 'scale' and 'auto' options. The default value is 'scale' 
\newline
\newline
After completing the grid search, the best parameters are extracted and the corresponding cross-validation accuracy. These tuned parameters are then used to re-train on the entire training set before evaluating on the test set in the next chapter. After exploring various hyperparameter combinations using GridSearchCV, the optimal SVM settings for our data are a regularization parameter C=10, gamma set to auto, and an RBF kernel. The tuned model achieved a cross-validation accuracy of\textbf{ 85.73\%}, indicating an that it is significantly better than the default settings. Basically, this means that with these new parameters, the SVM is better balanced between overfitting and underfitting, leading to improved performance when classifying vessels.

\subsection{Decision Tree Model}
Decision trees are chosen as the next step in the modeling phase, due to their ability to be able to capture non-linear relationships and interaction between features \cite{muller-2016}. Decision trees are supervised models that split recursively based on feature values to form what looks like a tree. At each node, the algorithm selects the best feature and threshold to sort the data into more homogeneous groups, aiming to reduce impurity using gini impurity/entropy \cite{perez-2019}. Key parameters in the model are max depth, which indicates the number of levels in the tree. A lower max depth helps prevent overfitting by restricting the model's complexity. Minimum samples split specifies the minimum number of samples required at a node to consider splitting it. This helps control overfitting by preventing the tree from creating splits based on too few samples.
\newline
\newline
Initially, a decision tree with its default parameters (max depth set to none, min samples split set to 2, min samples leaf set to 1 and with the gini impurity criterion) is modeled and evaluated using 5-fold cross-validation. This yielded a CV accuracy of \textbf{87.96\%} (\textbf{±2.38\%}), indicating that on average the tree correctly classifies nearly \textbf{88\%} of the instances with relatively low variability across folds.

\subsubsection{Decision Tree Model with hyperparameter tuning}
To improve the performance of the Decision Tree model with default parameters,  \cite{muller-2016}, hyperparameter tuning can help by fitting a better model. Using GridSearchCV, the decision tree parameters are tuned over a grid of possible values. As before, the accuracy of 5 fold cross-valuation accuracy are used. The optimal hyperparameters found were a maximum depth of 5, minimum sample split of 2, and minimum samples leaf split of 2 as well, which slightly improved the CV accuracy to \textbf{88.42\%}. As it turns out, tuning helps balance the tree’s complexity, leading to a larger capture for the essential patterns in the data without overfitting. Another option that could have been considered, is to perform pre-pruning, which involves restricting the growth of the tree by applying constraints like limiting the maximum depth of the tree, requiring a minimum number of points in a node to keep splitting it, or limiting the maximum number of leaves. This has not been done in this specific case.

\subsection{Random Forest Model}
Random Forest enhances the robustness previously described with Decision Trees by combining predictions from multiple trees, therefore improving accuracy and reducing overfitting. This also means that it is an ensemble learning method, since it builds multiple decision trees and outputs the mode (for classification) of their predictions. Thus it combines the predictions of many individual trees to (often) get better results \cite{provost-2013}. Ensamble means that instead of relying on a single decision tree which can be sensitive to overfitting and noise, RF builds many trees and aggregates their predictions. This is also known as bagging or bootstrap aggregation. Each tree is trained on a random subset (with replacement) of the training data \cite{muller-2016}. This initial RF model is done by using default hyperparameters. Afterwards hyperparameter tuning is performed. 

The key parameters in the RF model are as follows: 
\newline
\newline
\textbf{n\_estimators }: Defines the number of trees in the "forest", having an increase in the number of trees generally improves performance by reducing variance, but also increases the computational cost of estimating. The default value is 100.
\newline
\newline
\textbf{max\_depth }: Defines the max depth allowed for each tree, limiting the depth can prevent overfitting by ensuring that each tree does not expand to be too complex. The default value is none. 
\newline
\newline
\textbf{min\_samples\_split }: Defines the minimum number of samples required to split an internal node. This parameter controls how many samples must be present to allow a splitting to happen. Lower values allows the tree grow deeper, which might capture more detail but can lead to overfitting. The default value is 2.
\newline
\newline
When using the default parameters (e.g., the default number of trees and no specific cap on depth), the Random Forest was evaluated as previous models using cross-validation. Additionally, it uses bootstrap set to true, which means that each tree is trained on a random subset of data, helping with variance reduction. The default Random Forest already achieved a very high cross-validation accuracy of about \textbf{91.23\%} with very low variability, suggesting that the ensemble is both accurate and stable. Compared to the untuned Decision Tree (\textbf{87.96\%}\textbf{ ± 2.38\%}), the Random Forest significantly improves performance by reducing overfitting and variance through the ensemble method.

\subsubsection{Random Forest Model with hyperparameter tuning}
After tuning, the Random Forest achieved a slight improvement in accuracy to \textbf{91.30\%}. Although the gain is not as significant as other models after tuning, the tuned parameters help in handling the balance between complexity and generalization in the model. During hyperparameter tuning, for max depth one of the candidate values was None (i.e., unlimited) and specific depths such as 5, 10, and 20 were tested. The optimal hyperparameters were found to be \( n\_estimators = 200 \), \( max\_depth = 20 \), and \( min\_samples\_split = 2 \).
This fine-tuning, while only slightly boosting the accuracy compared to the default parameters, confirms that our Random Forest was already quote robust. The low variation (sd of around 0.0097) is maintained, indicating that the performance is consistent across different data splits. By averaging the predictions of multiple trees, Random Forest mitigates the risk of overfitting inherent in a single decision tree and as is proven by the significant change in accuracy, this builds a more complex but more accurate model.

\subsection{Class imbalance}
As described above, SMOTE (Synthetic Minority Over-sampling Technique) \cite{chawla-2002} is a method used to address class imbalance in the dataset by generating synthetic data examples for the minority classes rather than duplicating existing samples. It creates new, plausible instances by interpolating between existing minority examples \cite{muller-2016}. This results in a more balanced training dataset, allowing models to learn more robust decision boundaries and improving the recognition of underrepresented classes. In this case it helps with ship types that are not as present in the dataset within the specific time range, with the smallest class having only 48 instances in the training data. After applying SMOTE augmented (balanced) training data, each class in the training data has 799 samples. This synthetic balancing changes the dynamics of model training. GaussianNB, SVM, Decision Tree, and Random Forest all show higher cross-validation accuracies (Random Forest CV accuracy on the training set for example increases from \textbf{91.2}\% to ~\textbf{97.2}\%). This boost is expected because the model sees a balanced representation of all classes during training. The GaussianNB model's performance is relatively unaffected by the use of SMOTE indicating a limited ability to effecticely use balanced training data in the context of this ship type detection task. Further insights are made in the next section by reviewing table \ref{tab:classifier_performance}

\subsection{Conclusion}
In this section, 5 different models were fitted and evaluated with 5-fold cross-validation on the training data. Afterwards, the models' hyperparameters were tuned according to the cross-validation accuracy. At last, SMOTE was applied to handle case imbalance, and models were hyperparameter tuned again. It was found that SMOTE dramatically improves CV scores because it balances the training set.  It was found that GaussianNB consistently ranks as the lowest performer, while SVM ranks a bit higher, but still rank lower than the tree-based models like DT and RF. The hyperparameter tuning with grid search improved CV accuracy for SVM, Decision Tree and Random Forest both with and without SMOTE applied. In the next section, all models will be evaluated on the test set, and feature importance as well as error analysis will be performed. 

\section{Results}
In this section, the performance of the classification models is analyzed and evaluated to better understand their capabilities and limitations when applied in practice.
\newline
\subsection{Classifier model results}
\begin{table}[H]
    \centering
    \begin{tabular}{l|c|c|c|c}
        \textbf{Classifier} & \textbf{Accuracy \%} & \textbf{Precision \%} & \textbf{Recall \%} & \textbf{Avg. F1 (\%)}  \\
        \hline
        GaussianNB (Default) & 76.70\% & 76.98\% & 85.63\% & 79.93\%  \\
        SVM (Default) & 82.46\% & 89.01\% & 78.44\% & 78.80\%  \\
        Decision Tree (Default) & 88.74\% & 91.29\% & 90.62\% & 90.88\%  \\
        Random Forest (Default) & 90.84\% & 94.46\% & 88.46\% & 90.91\%  \\
        SVM Tuned & 85.08\% & 90.75\% & 84.92\% & 86.12\%  \\
        Decision Tree Tuned & 89.27\% & 92.24\% & 87.49\% & 89.48\%  \\
        Random Forest Tuned & 91.36\% & \textbf{94.95\%} & 90.43\% & 92.24\%  \\
        GaussianNB (Default + SMOTE) & 76.44\% & 77.52\% & 86.06\% & 80.17\%  \\
        SVM (Default + SMOTE) & 77.75\% & 83.44\% & 87.28\% & 84.31\%  \\
        Decision Tree (Default + SMOTE) & 88.48\% & 91.09\% & 90.97\% & 90.92\%  \\
        Random Forest (Default + SMOTE) & \textbf{92.15\%} & 94.11\% & \textbf{92.51\%} & \textbf{93.27}\%  \\
        SVM Tuned + SMOTE & 84.56\% & 88.32\% & 89.49\% & 88.70\%  \\
        Decision Tree Tuned + SMOTE & 88.48\% & 91.09\% & 90.97\% & 90.92\%  \\
        Random Forest Tuned + SMOTE & 91.88\% & 93.96\% & 92.21\% & 93.04\%  \\
    \end{tabular}
    \caption{Summary of classifier performance on test set (values in \%)}
    \label{tab:classifier_performance}
\end{table}
The original imbalanced training data leads to weak accuracies, especially for SVM and GaussianNB—in accurately predicting minority classes (specifically the tanker class, as visible in the confusion matrix in appendix). For GaussianNB, when looking at class 4 (Tanker), precision is only 0.49 and recall is 0.79, resulting in an F1 score of 0.60. While SMV shows good overall accuracy with default settings, its performance on minority classes is poor. Tuning helps improve the overall metrics a bit, but it still struggles with certain classes. Random Forest, especially, consistently achieves the highest accuracy and balanced precision/recall/F1 scores. There is a hint, however, of potential overfitting in that the default Random Forest performs nearly as well as its hyperparameter-tuned counterpart, with only marginal improvements seen after tuning. This small gap between default and tuned performance suggests that the model might already be generalizing well, and further tuning could risk fitting too closely to the training data rather than improving generalization to unseen data. This suspicion is strengthened when looking at the test set classification accuracy, where the default Random Forest with SMOTE outperforms the tuned Random Forest on every benchmark except precision.

\subsubsection{Error analysis}
To identify which classes have weak predictions, a confusion matrix is done. A confusion matrix provides a clear picture of how often each class is misclassified, highlighting patterns of errors. The confusion matrix in full size can be found in Appendix 12. The confusion matrix highlights that class 1 has a significant number of errors. Specifically, it is often misclassified as class 0 on 15 samples. This suggests that the features for class 1 ships may overlap with those of class 0, or the model is having difficulty separating these two classes. This is confirmed by seeing that 11 samples where class 1 is misclassified as class 0. By manually inspecting those classes, we can see that class 0 is the cargo ship type, where class 1 is the tanker ship type. This makes sense, as these vessels often exhibit the same trajectories, merely passing through the Baltic Sea. To help in this classification, additional features that better help seperate these two classes could be included. ROC AUC scores are calculated for three tuned classifiers (SVM/RF/DT tuned), computed using a one-vs-rest (OvR) approach. The values are: SVM Tuned: \textbf{0.9477}, Decision Tree Tuned: \textbf{0.9614}, Random Forest Tuned: \textbf{0.9897}. All three models perform well (with values above \textbf{0.94}), but the Random Forest Tuned model shows the highest ROC AUC, suggesting that it is best at distinguishing between ship types across the board.

\subsubsection{Feature importance}
\begin{figure}[H]
    \centering
    \begin{subfigure}{0.48\linewidth}
        \centering
        \includegraphics[width=\linewidth]{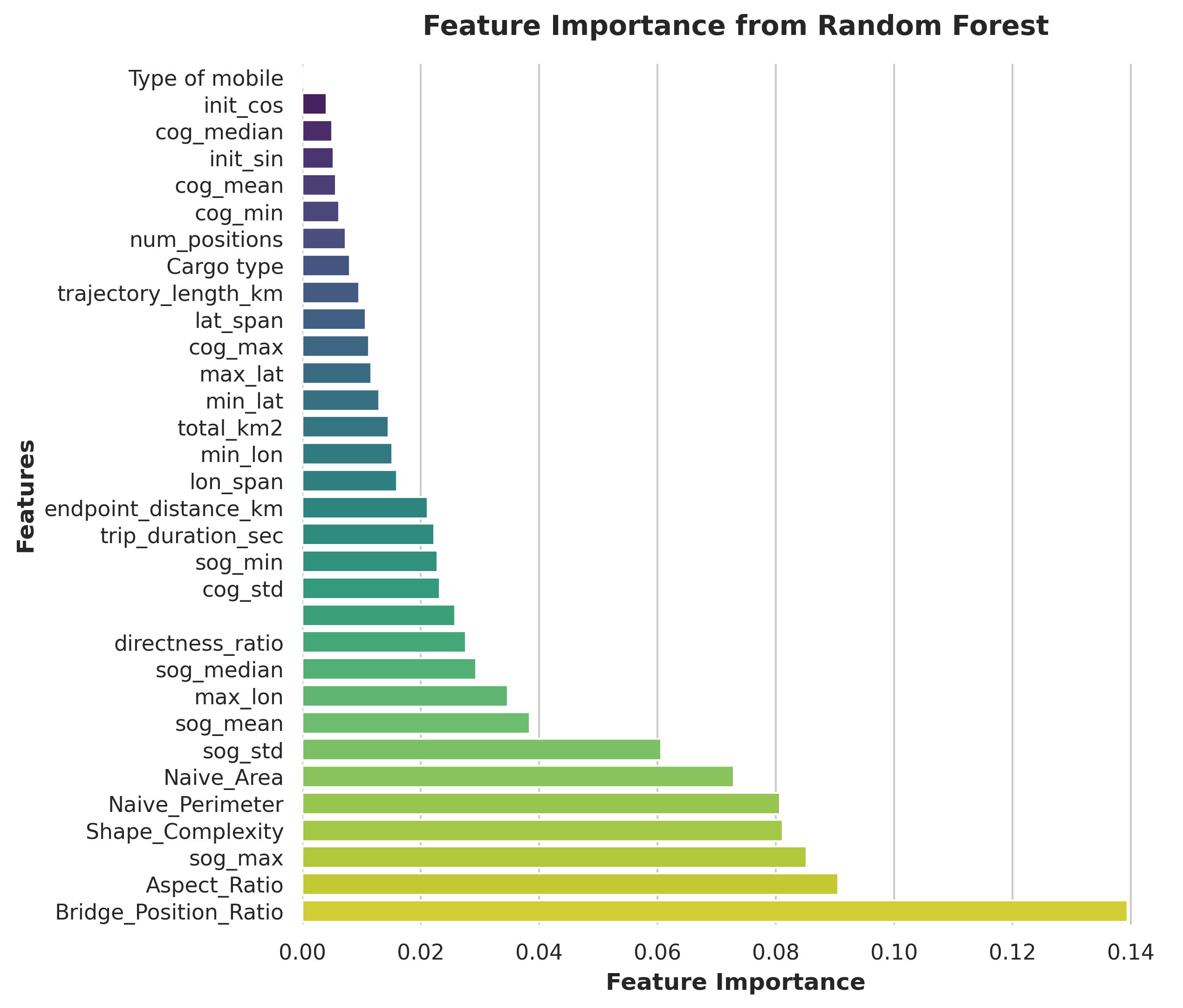}
        \caption{Feature importance from Random Forest Tuned + SMOTE}
        \label{fig:featureimportance1}
    \end{subfigure}
    \hfill
    \begin{subfigure}{0.48\linewidth}
        \centering
        \includegraphics[width=\linewidth]{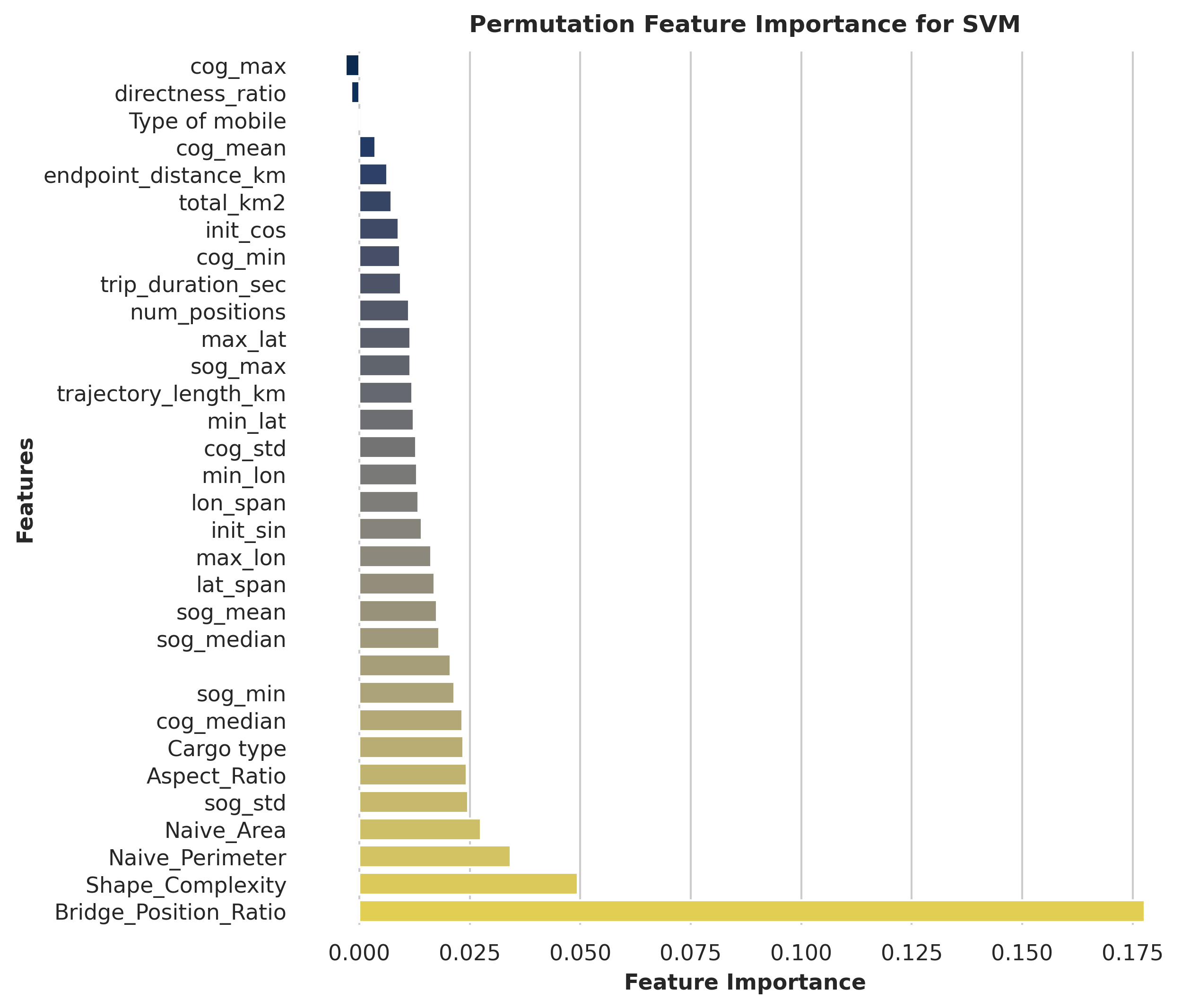}
        \caption{Permutation feature importance Tuned SVM + SMOTE}
        \label{fig:featureimportance2}
    \end{subfigure}
    \caption{Comparison of feature importance from RF/SVM}
    \label{fig:feature_importance_combined}
\end{figure}

\begin{figure}[H]
    \centering
    \includegraphics[width=0.5\linewidth]{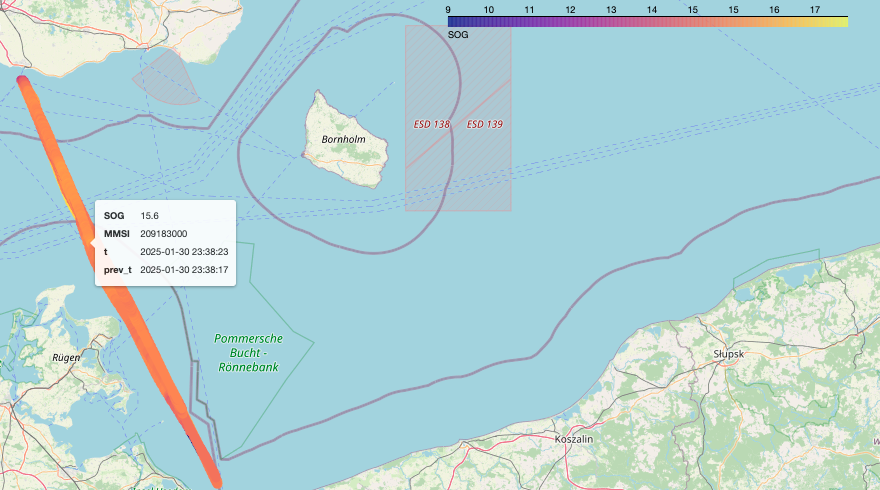}
    \caption{Failure to segment trajectory, ferry being segmented into one trip from 06:10:03 -> 21:55:09 although it is clearly different trips with 45 minute stops in each port}
    \label{fig:shippolandsweden}
\end{figure}

To better understand the background of the well performing Random Forest model, a feature importance analysis is done. Feature importance covers the importance of each feature when it comes to the decision a tree makes \cite{muller-2016}.  The Random Forest model is an explainable machine learning model, which means that it requires post-hoc techniques to explain and ensure its understandability and interpretability \cite{chawla-2002}. While a Random Forest model is rather advanced compared to more simple models, doing feature importance analysis increases the transparency of the model. Transparency becomes an important matter when applying a machine learning model on real-world scenarios and when used for actual decision making. For example, if a government entity utilizes the classification model to consider whether to send out helicopters to learn more about a specific ship based on it's type classification (for example a fishing ship type that is being classified as a passenger ship), it could have large side effects in that resources are perhaps being allocated in a wrong way. Another 'worst case' scenario, is if an innocent ship operator came under suspicion due to the outcome of a ship type machine learning classification model. By increasing transparency for a model, this could help prevent misunderstanding of the models predictions, as operators relying on the ML classifications would be able to better understand what real-world factors help drive specific end results. The importance values are derived from how much a feature reduces impurity (measured by Gini impurity or entropy, in this case Gini impurity) across all the decision trees in the Random Forest. Features that frequently help split data into pure groups get higher importance scores, while those that contribute little get lower scores. The score visible in figure \ref{fig:featureimportance1} for each feature is calculated by summing the impurity reduction across all trees where it was used for a split.
When reviewing figure \ref{fig:featureimportance1} it is visible that the most influential feature is the bridge position ratio, explained in the feature extraction chapter. This likely plays a key role in the model’s decision-making, this makes since, since it is directly related to what type of ship are being transmitted in the AIS messages. Maximum Speed over Ground (SOG) is also an important factor in predictions. It makes sense, since vessels have a specific top speed, and a cargo ship can never reach the same top speeds as a passenger ship for example. SOG standard deviation, SOG mean, SOG median (0.06-0.03) contribute to the model but are not dominant factors. Trip duration in seconds also contributes, but less so than shape-based features and SOG.
\newline
\newline
To sum it up, the models heavily relies on spatial and geometric properties rather than properties like cargo type or trajectory length. The fact that trajectory length, endpoint distance, trajectory time are not very significant could indicate that the trajectory segmentation are not as clear cut as possible. This worry is strengthened when manually reviewing one of the classifications, a passenger ferry ship (MMSI: 209183000) on route "\textit{Swinoujscie to Trelleborg}". As can be seen on figure \ref{fig:shippolandsweden}, the ferry routes are segmented into one continuous trajectory, since the ferry only docks for less than the threshold and onboards new passengers. This severely distorts the trajectory specific variables in the dataset. It is worth noting, that they are further distorted due to the rather small Area of Interest. All tankers/cargo ships are passing through, meaning they get the exact same distance in nautical miles, and having worldwide AIS traffic data could find mean better trajectory features for classifications. Overall, vessel specific values is the most important for classification, and it could be helpful to further research how to better segment continuous AIS messages into correct trajectories or trips.
\section{Discussion and further work}
Throughout the study, various inefficiencies and alternative approaches are mentioned in passing. In this chapter, these points are discussed to consider how the classification models could be improved even more.
\begin{enumerate}[left=0pt]
    \item \noindent \textbf{Decision tree models perform well, what about XGBoost?}
\newline
\noindent
Other research has been conducted on this topic \cite{wang-2021}, \cite{wu-2016}, \cite{ferreira-2023}, \cite{huang-2023}, \cite{wang-2022}, and the accuracy of the Random Forest model as well as the Decision Trees is on-par with what is found in \cite{huang-2023}, where a comparable strait area between two Taiwanese wind farms was used, albeit with a longer time scope. In the 2023 study, XGBoost was also used, and performed better than RF and DT. Had the scope of the paper allowed it, it could have been interesting to see if XGBoost could outperform the models used in this study in Danish waters. Another study from Texas A\&M University on AIS data from the danish waters trained a Random Forest classification model with 84\% average accuracy \cite{baeg-2024}. The main difference however, is they in this study only a subset of data from a specific strait in Bornholm is used, where mostly cargo/tanker/ vessels are passing through. In the other study, country-wide AIS traffic was used, which causes severe disturbances from vessels operating in ports, close to shore, close to wind farms, etc. which obscures the natural speed/turning features of the ship, that we utilized here.
    \item \noindent \textbf{Inaccurate trajectories}
\newline
\noindent As found in the feature importance analysis section above, trajectory features are not as important as vessel details in both the Random Forest and Support Vector classifier, reason being that trajectory segmentation sometimes fail due to the rather simple rules put in place to segment trips. By further experimenting with alternative approaches to pre-processing the dataset, the classifiers could be improved. A suggestion could be to implement an unsupervised spatial clustering algorithm like DBSCAN to figure out where ships usually anchor/start/stop trajectories and split trips there. 
    \item \noindent \textbf{Moving vessel detection to static vessel detection}
\newline
\noindent Since the polygon bounds are all drawn to filter AIS traffic to only cover at least 0.5 nautical miles outside of ports and shoreline, no data is trained upon where vessels are still. This was done to make it easier to classify \textit{moving} vessels. It could be argued that it is just a matter of training the correct model. By bringing in all of the AIS data within Danish waters as well as all ports and on land, it could be that models would perform even better, and also be able to classify more static vessels.  
\end{enumerate}

\noindent By utilizing the trained tuned \textbf{RF} (w/ SMOTE) classification model approach on a new dataset for 31. Jan 2025, but only applying the classification for messages where ship types are missing, we can see the clear operational value of the model and approach. A snippet from this data is shown in figure \ref{fig:backfilled-ml-detect} below. The worlds largest provider of AIS traffic, \textit{MarineTraffic}, charges north of 200\$ for a subscription that (among other features) allows people to view "\textit{AI detected}" vessel types for Satellite AIS reported vessels. This shows the potential impact and value of publishing this sort of research openly. 
\begin{figure}[H]
    \centering
    \includegraphics[width=0.75\linewidth]{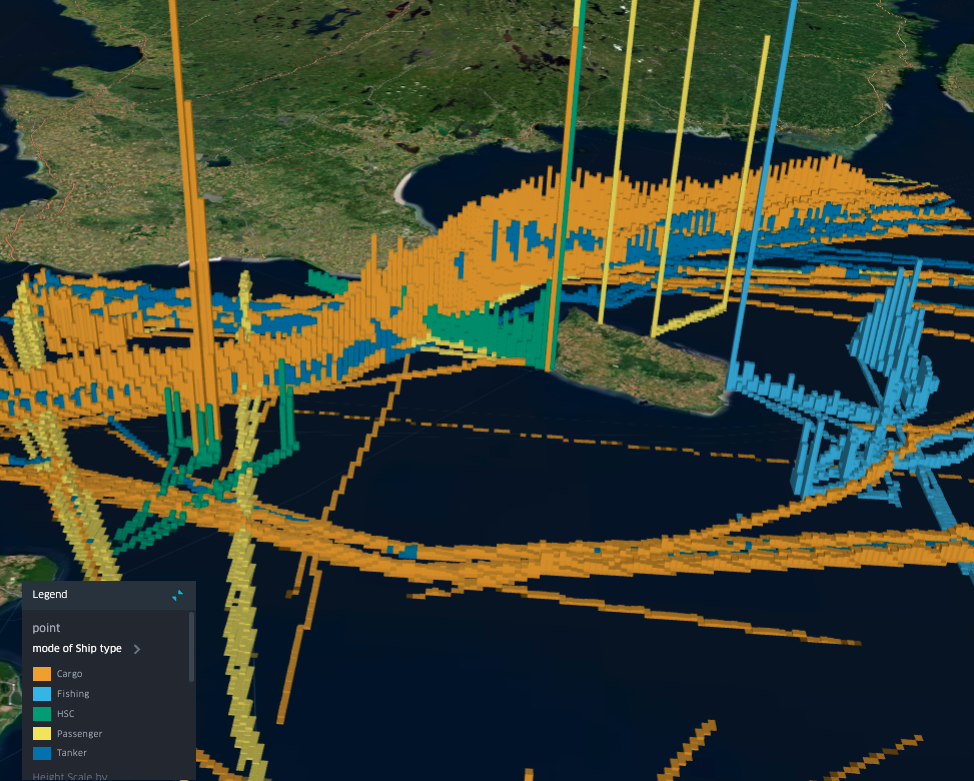}
    \caption{31 Jan 2025 - Random forest detected ship types for missing ship types in AIS messages (backfilled/forwardfilled based on MMSI)}
    \label{fig:backfilled-ml-detect}
\end{figure}
\newpage
\section{Conclusion}
In this study, an approach was presented for the problem of classifying vessel types based on moving vessel AIS data for a specific area. For this purpose, movements of the trajectory based on the dynamic information of latitude and longitude as well as ship and voyage related information are used. Initially, the AIS messages were preprocessed and segmented based on tracks, with features being extracted into a collective trip covering many AIS messages. In sequence, various classifications models were applied. By evaluating the models using various benchmarks, the RF model with SMOTE applied to the dataset showed to be the best performing model on the test set with an accuracy of around 92\%. Utilizing the fitted Random Forest classifier on completely new data, not seen in train data or test data, makes it possible to identify ships cruising around in the Baltic Sea with great accuracy as seen below.
\begin{figure}[H]
    \begin{subfigure}[b]{0.5\linewidth}
        \centering
        \includegraphics[width=\linewidth]{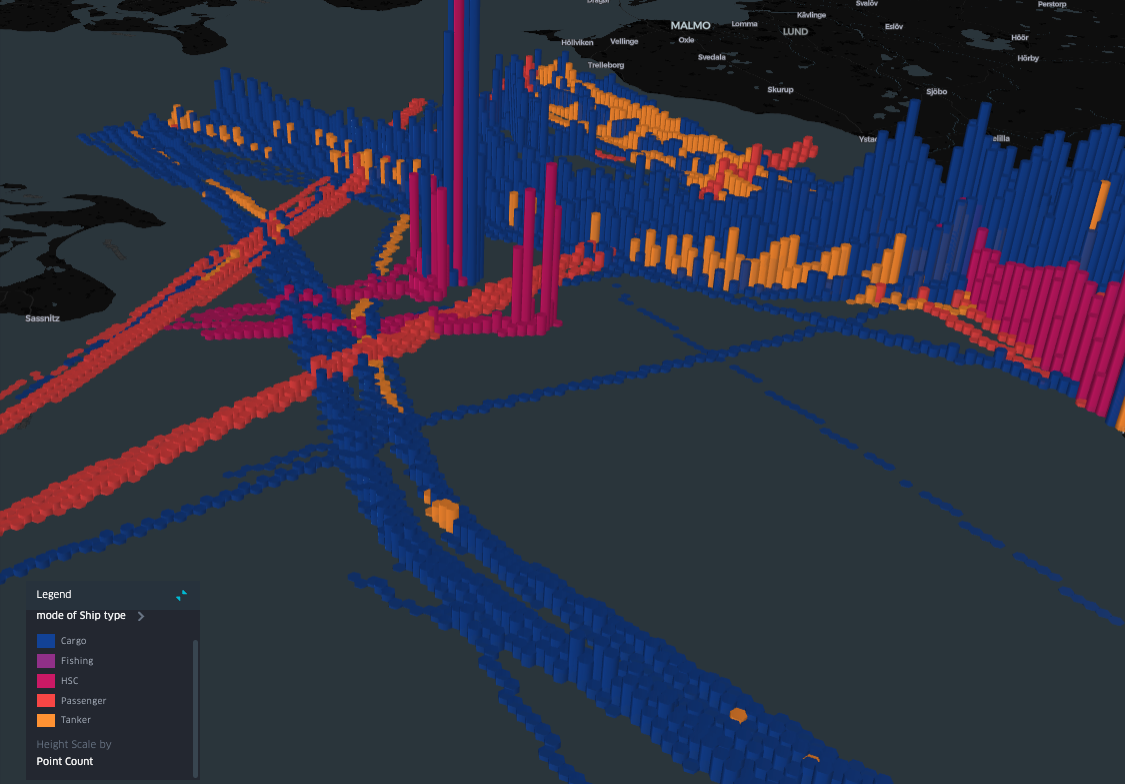}
        \caption{Backfilling classification model on AIS transmissions with missing Ship Types (Visualized with kepler.gl)}
        \label{fig:balticsea2}
    \end{subfigure}
    \begin{subfigure}[b]{0.5\linewidth}
        \centering
        \includegraphics[width=\linewidth]{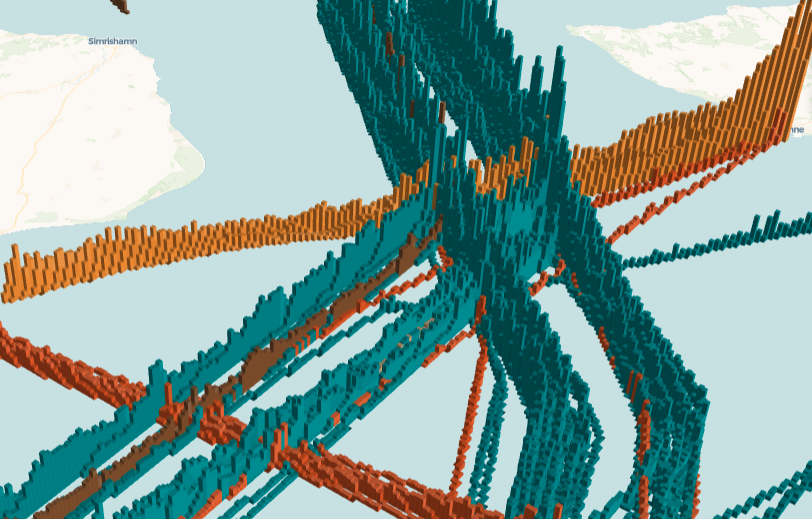}
        \caption{Detected cruise routes, HSC routes, and passenger routes (Visualized with kepler.gl)}
        \label{fig:balticsea1}
    \end{subfigure}
    \caption{Comparison of backfilling classification model and thereby detecting ship type trajectories.}
    \label{fig:comparison}
\end{figure}
\newpage
\bibliographystyle{apalike}
\bibliography{sample}
\newpage
\section{Appendix}
\appendix
\subsection{Polygon bounds}
[ [ 14.270042565757286, 55.43261413541811 ], [ 14.159660553673517, 55.31431394562776 ], [ 13.943485869589487, 55.37509769839545 ], [ 13.736216591081442, 55.38441331424108 ], [ 13.541496069513602, 55.35068289739041 ], [ 13.304762600364848, 55.3120526417598 ], [ 12.890616626826521, 55.35737054573761 ], [ 12.560203703925447, 55.29416763469967 ], [ 12.386701794787081, 55.11720877415865 ], [ 12.603693008332394, 54.969233301575066 ], [ 12.414218417997406, 54.820638173689154 ], [ 12.680196098521764, 54.71037433774884 ], [ 13.079986989396057, 54.62303076992553 ], [ 13.468249950804049, 54.791020033388484 ], [ 13.51170971146399, 54.59832703857482 ], [ 13.737123206805933, 54.599532023328116 ], [ 13.726267920343783, 54.47789437636281 ], [ 13.80248977691796, 54.44431060639897 ], [ 13.816069958005013, 54.25320187474711 ], [ 14.00875608215033, 54.099696784013666 ], [ 14.374713839968468, 53.9970757396895 ], [ 14.493199683979855, 54.01043871434795 ], [ 14.69316702656555, 54.039715072849376 ], [ 15.0716126364171, 54.15181830569306 ], [ 15.307556101054333, 54.19829090233562 ], [ 15.433737254059501, 54.219319534665914 ], [ 15.587213933721209, 54.231170229650914 ], [ 15.760706246048885, 54.261504194731344 ], [ 15.893325254169469, 54.29035882936187 ], [ 16.088906570305276, 54.35665628585242 ], [ 16.129551606193363, 54.43229272091137 ], [ 16.07378248639351, 54.506990798416574 ], [ 16.071707542331847, 54.6207919243221 ], [ 16.151084277320372, 54.703276474917025 ], [ 16.316183574856048, 54.878985427323144 ], [ 16.185762815825427, 55.032487511103405 ], [ 16.022249754316388, 55.263159654985714 ], [ 14.94633767765143, 55.24975322797314 ], [ 14.989481884064402, 55.23175499690595 ], [ 15.057754966863946, 55.194093599183475 ], [ 15.149476602505743, 55.17775630670721 ], [ 15.186039541704602, 55.10658148747229 ], [ 15.169752683294195, 55.03007704275702 ], [ 15.08403056247248, 54.96996354072045 ], [ 14.886938417059259, 54.99971171035721 ], [ 14.70836384348897, 55.051135527462726 ], [ 14.659047407611874, 55.126136026081895 ], [ 14.680956989048685, 55.229802490475606 ], [ 14.769200731909004, 55.318211974097125 ], [ 15.053715508170237, 55.22903165438766 ], [ 16.056195026446222, 55.246488932310626 ], [ 16.040908236572225, 55.66958011622285 ], [ 16.172181461304735, 55.87160815960177 ], [ 16.055258173762706, 55.88783732919536 ], [ 15.89946307176809, 55.90960258872895 ], [ 15.735262190453325, 55.91288018401511 ], [ 15.629949715260613, 55.901890079143655 ], [ 15.528182723935682, 55.91158064472876 ], [ 15.325246448324696, 55.914561883012205 ], [ 15.20730036968636, 55.91768618110561 ], [ 14.982694397745743, 55.92741101403853 ], [ 14.790771706957553, 55.94381732572509 ], [ 14.63178811498484, 55.94202233031005 ], [ 14.514148681878968, 55.932482624665084 ], [ 14.42025881204937, 55.90693635359379 ], [ 14.345549271550066, 55.82838664780491 ], [ 14.294572271785078, 55.76246662446725 ], [ 14.323345581638538, 55.69540583895088 ], [ 14.362663965300833, 55.6414674327225 ], [ 14.372349875380577, 55.629759612922975 ], [ 14.392639023006781, 55.61049134061141 ], [ 14.419946988454956, 55.5843963369995 ], [ 14.42082516398884, 55.56703906821493 ], [ 14.413903860207025, 55.54595071001545 ], [ 14.4010070379913, 55.515193046204615 ], [ 14.362030314861514, 55.47991837055746 ], [ 14.346057981017491, 55.465635294498604 ], [ 14.323402126778861, 55.45243414314477 ], [ 14.297726058444235, 55.44187815084099 ], [ 14.270042565757286, 55.43261413541811 ] ]
\subsection{Histograms for key numerical features}

\begin{figure}[H]
    \centering
    \includegraphics[width=1\linewidth]{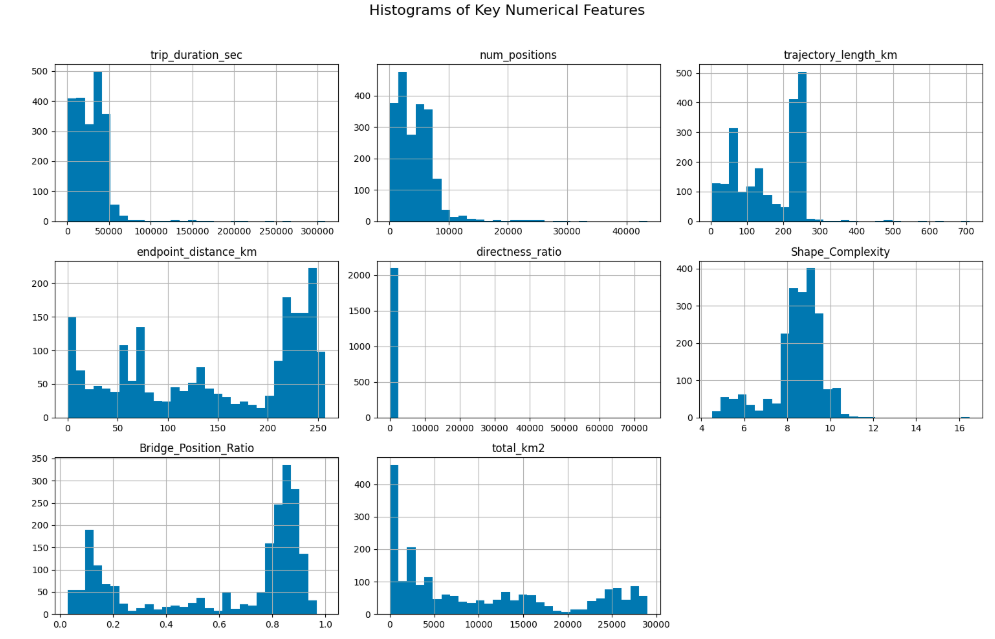}
    \caption{Feature histograms}
    \label{fig:featurehisograms}
\end{figure}

\subsection{Correlation matrix of selected features}

\begin{figure}[H]
    \centering
    \includegraphics[width=1\linewidth]{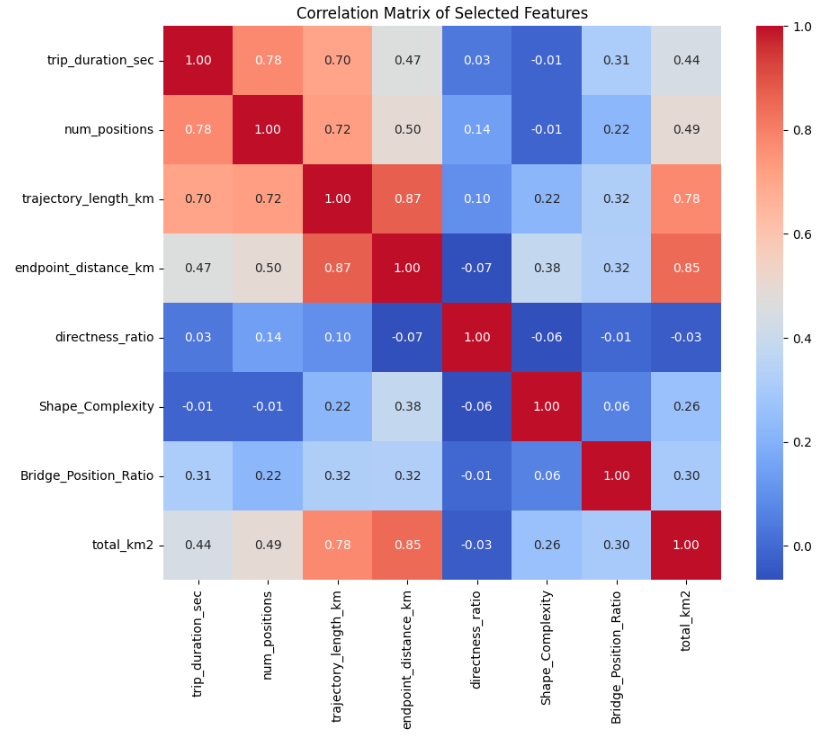}
    \caption{Matrix for features}
    \label{fig:cormatrix}
\end{figure}

\subsection{Trip duration by ship type}
\begin{figure}[H]
    \centering
    \includegraphics[width=1\linewidth]{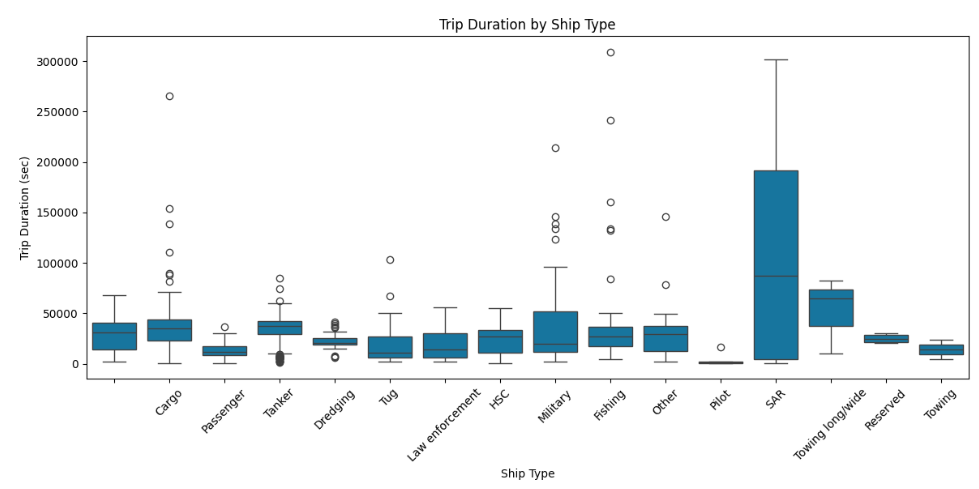}
    \caption{Ship type and trip duration}
    \label{fig:shiptypeduration}
\end{figure}

\subsection{Trajectory length and trip duration}
\begin{figure}[H]
    \centering
    \includegraphics[width=1\linewidth]{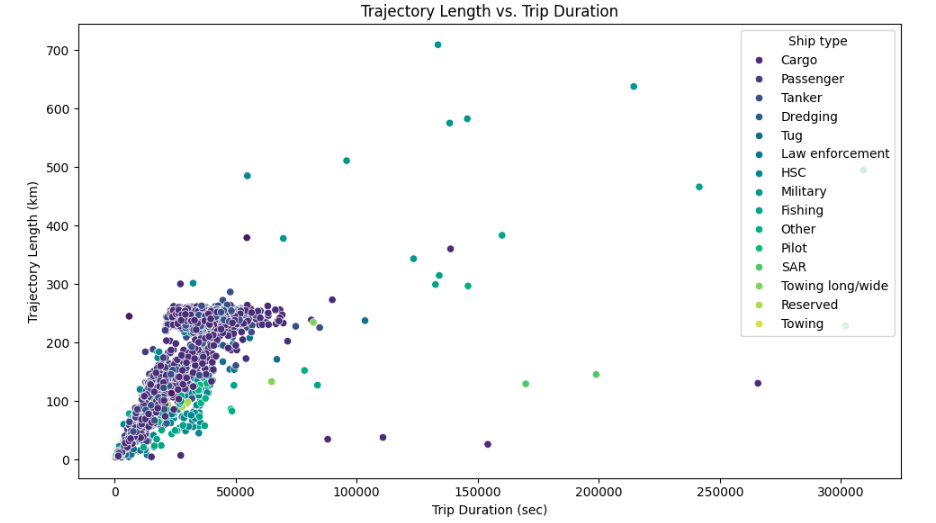}
    \caption{Trajectory length and trip duration}
    \label{fig:trajlengthandtripduration}
\end{figure}

\subsection{Pairplot of selected features}
\begin{figure}[H]
    \centering
    \includegraphics[width=1\linewidth]{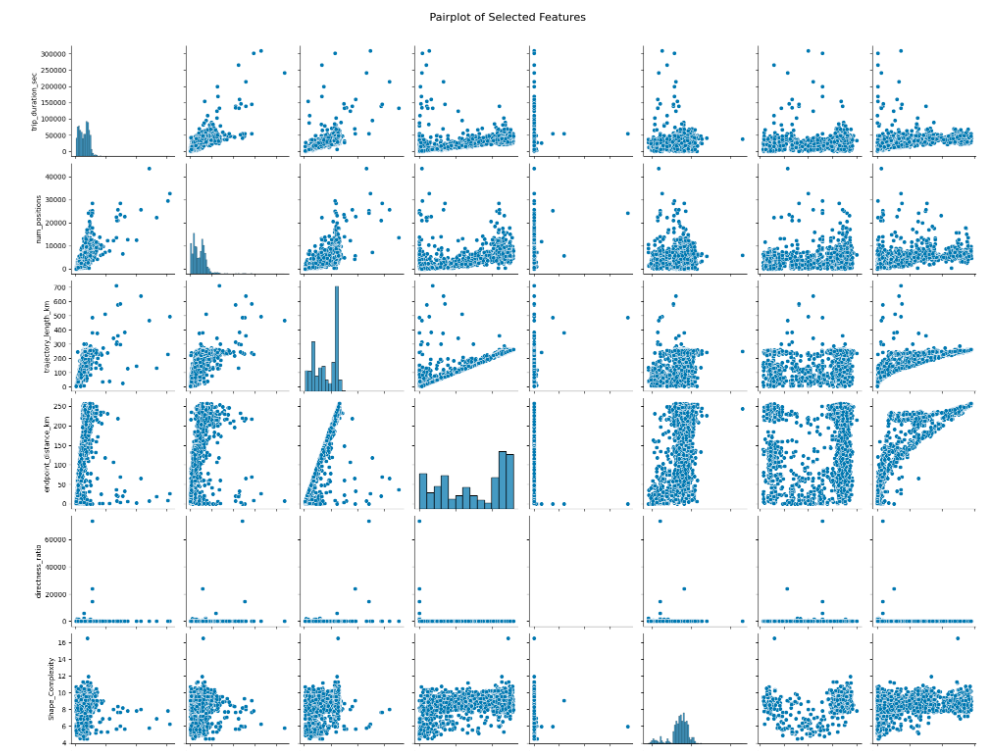}
    \caption{Pairplot}
    \label{fig:Pairplot}
\end{figure}

\subsection{Accuracy, precision, recall, F1 formulas}
True Positives (TP), True Negatives (TN), False Positives (FP), and False Negatives (FN) are defined as follows:

\begin{itemize}
    \item TP: The number of correctly classified positive samples.
    \item TN: The number of correctly classified negative samples.
    \item FP: The number of negative samples incorrectly classified as positive.
    \item FN: The number of positive samples incorrectly classified as negative.
\end{itemize}

The total number of samples is:

\[
N = TP + TN + FP + FN
\]

The accuracy is given by:

\[
\text{Accuracy} = \frac{TP + TN}{TP + TN + FP + FN}
\]

Precision (also called Positive Predictive Value) is given by:

\[
\text{Precision} = \frac{TP}{TP + FP}
\]

Recall (also called Sensitivity or True Positive Rate) is given by:

\[
\text{Recall} = \frac{TP}{TP + FN}
\]

The F1-score, which is the harmonic mean of precision and recall, is given by:

\[
F1 = 2 \times \frac{\text{Precision} \times \text{Recall}}{\text{Precision} + \text{Recall}}
\]
\subsection{Military ships in the baltic sea in dataset}
\begin{figure}[H]
    \centering
    \includegraphics[width=0.5\linewidth]{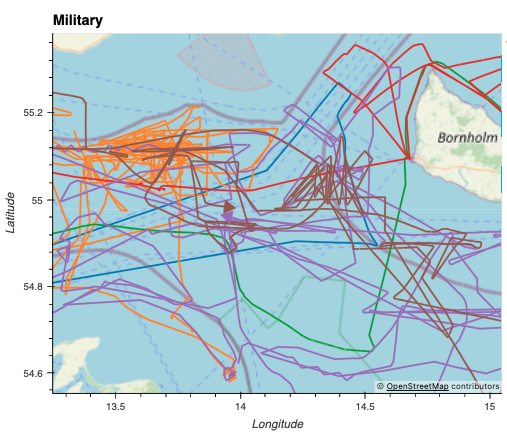}
    \caption{Military ships in dataset around bornholm}
    \label{fig:mil}
\end{figure}
\subsection{All rows in raw AIS data}
\begin{table}[H]
    \small
    \centering
    \begin{tabular}{l|l|l}
        \textbf{Name} & \textbf{Description} & \textbf{msg type}\\ \hline
        Timestamp & The exact date and time of the data entry. & Dynamic\\ \hline
        Mobiletype& The classification of the AIS mobile device in use.& Dynamic\\ \hline
        MMSI & Maritime Mobile Service Identity, a unique identifier for a vessel. & Dynamic\\ \hline
        Latitude & The latitude coordinate of the vessel's position. & Dynamic\\ \hline
        Longitude & The longitude coordinate of the vessel's position. & Dynamic\\ \hline
        Nav status& The current navigational status of the vessel. & Dynamic\\ \hline
        ROT & Rate of Turn, the rate at which the vessel is turning. & Dynamic\\ \hline
        SOG & Speed Over Ground, the speed of the vessel relative to the ground. & Dynamic\\ \hline
        COG & Course Over Ground, the direction of the vessel's path. & Dynamic\\ \hline
        Heading & The actual compass direction the vessel is heading. & Dynamic\\ \hline
        IMO & International Maritime Organization number, a unique vessel identifier. & Static\\ \hline
        Callsign & The call sign of the vessel. & Static\\ \hline
        Name & The name of the vessel. & Static\\ \hline
        Ship type & The classification of the ship. & Static\\ \hline
        Cargo type & The type of cargo the ship is carrying. & Static\\ \hline
        Width & The width of the vessel. & Static\\ \hline
        Length & The length of the vessel. & Static\\ \hline
        Pos fix device& The device used for positioning the vessel. & Static\\ \hline
        Draught & The depth of water needed to float the vessel. & Static\\ \hline
        Destination & The intended destination of the vessel. & Static\\ \hline
        ETA & Estimated Time of Arrival at the destination. & Static\\ \hline
        Data source& The origin of the data source. & Static\\ \hline
        A & (9 bits) Distance from bow to reference point& Static\\ \hline
        B & (9 bits) Distance from stern to reference point& Static\\ \hline
        C & (6 bits) Distance from port side to reference point& Static\\ \hline
        D & (6 bits) Distance from starboard side to reference point& Static
    \end{tabular}
    \caption{Table showing the name, description, and type of data fields related to maritime vessels.}
    \label{tab:maritime_data_fields}
\end{table}

\subsection{Stratified k-Fold Cross-Validation}
\begin{itemize}
    \item Split dataset into \( k \) folds while preserving class proportions:
    \[
    P(y)_{\text{train}}^{(i)} \approx P(y)_{\text{val}}^{(i)}
    \]
    \item Compute performance metric for each fold:
    \[
    M^{(i)} = f(\mathcal{D}_{train}^{(i)}, \mathcal{D}_{val}^{(i)})
    \]
    \item Average across all folds:
    \[
    M_{\text{cv}} = \frac{1}{k} \sum_{i=1}^{k} M^{(i)}
    \]
\end{itemize}

\subsection{Full list of Features by Feature, Type, Description}
\begin{longtable}{|l|l|l|}
    \hline
    \textbf{Feature} & \textbf{Type} & \textbf{Description} \\
    \hline
    \endfirsthead

    \hline
    \textbf{Feature} & \textbf{Type} & \textbf{Description} \\
    \hline
    \endhead

    \hline
    \multicolumn{3}{r}{\textit{Continued on next page...}} \\ 
    \hline
    \endfoot

    \hline
    \endlastfoot
    \hline
    MMSI & int64 & Maritime Mobile Service Identity, a unique identifier for the ship. \\
    Trip ID & int64 & A unique identifier for the trip. \\
    Trip Start & datetime64[s] & The timestamp when the trip started. \\
    Trip End & datetime64[s] & The timestamp when the trip ended. \\
    Ship Type & object & The category of the ship (e.g., cargo, tanker, fishing vessel). \\
    Cargo Type & object & The type of cargo the ship carries. \\
    Callsign & object & The radio callsign of the ship. \\
    Name & object & The name of the ship. \\
    Destination & object & The ship's intended destination. \\
    Trip Duration Sec & float64 & The total duration of the trip in seconds. \\
    Number of Positions & int64 & The number of AIS position reports received during the trip. \\
    Trajectory Length KM & float64 & The total distance traveled by the ship in kilometers. \\
    Endpoint Distance KM & float64 & The straight-line distance between the start and end points of the trip. \\
    Directness Ratio & int64 & Ratio of the straight-line distance to the actual trajectory length. \\
    Min Lat & float64 & Minimum latitude recorded during the trip. \\
    Max Lat & float64 & Maximum latitude recorded during the trip. \\
    Min Lon & float64 & Minimum longitude recorded during the trip. \\
    Max Lon & float64 & Maximum longitude recorded during the trip. \\
    Lat Span & float64 & The difference between max and min latitude. \\
    Lon Span & float64 & The difference between max and min longitude. \\
    SOG Min & float64 & Minimum Speed Over Ground (SOG) recorded during the trip. \\
    SOG Max & float64 & Maximum Speed Over Ground (SOG) recorded during the trip. \\
    SOG Mean & float64 & Mean Speed Over Ground (SOG) over the trip. \\
    SOG Median & float64 & Median Speed Over Ground (SOG) over the trip. \\
    SOG Std & float64 & Standard deviation of Speed Over Ground (SOG). \\
    COG Min & float64 & Minimum Course Over Ground (COG) recorded. \\
    COG Max & float64 & Maximum Course Over Ground (COG) recorded. \\
    COG Mean & float64 & Mean Course Over Ground (COG) during the trip. \\
    COG Median & float64 & Median Course Over Ground (COG) during the trip. \\
    COG Std & float64 & Standard deviation of Course Over Ground (COG). \\
    Init Cos & float64 & Cosine of the initial course angle. \\
    Init Sin & float64 & Sine of the initial course angle. \\
    Naive Perimeter & float64 & The perimeter of the ship's trajectory bounding box. \\
    Naive Area & float64 & The area of the ship's trajectory bounding box. \\
    Aspect Ratio & float64 & The aspect ratio of the bounding box (width/height). \\
    Shape Complexity & float64 & A measure of the complexity of the ship's movement pattern. \\
    Bridge Position Ratio & float64 & The relative position of the ship’s bridge within its trajectory. \\
    Type of Mobile & object & The classification of the mobile station (e.g., vessel, aid-to-navigation). \\
    Total KM2 & float64 & The total area covered by the ship during its trip. \\
    \hline
\end{longtable}

\subsection{Confusion matrix}
\begin{figure}[H]
    \centering
    \includegraphics[width=1\linewidth]{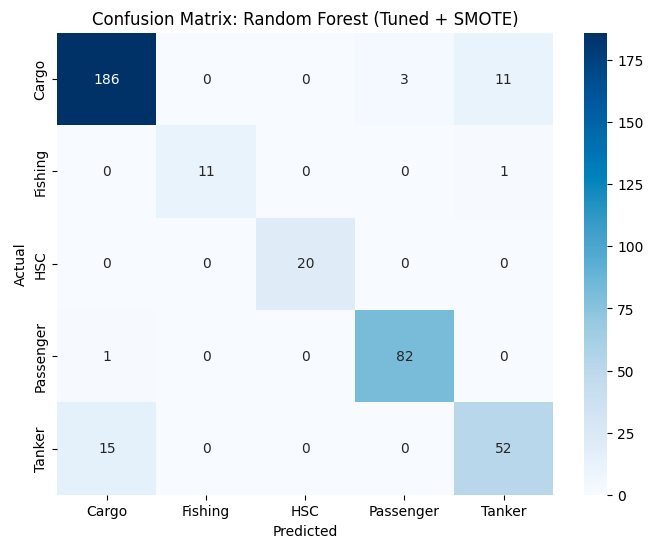}
    \caption{Confusion matrix}
    \label{fig:confusionmasxtr}
\end{figure}
\end{document}